%% file: main_cam.tex
\pdfoutput=1

\documentclass[11pt]{article}

\usepackage[final]{acl}

\usepackage{times}
\usepackage{latexsym}

\usepackage[T1]{fontenc}

\usepackage[utf8]{inputenc}

\usepackage{microtype}

\usepackage{inconsolata}

\usepackage{amsmath}
\usepackage{blindtext}
\usepackage{graphicx}
\usepackage[normalem]{ulem}
\usepackage{tabularx}
\usepackage{multicol}
\usepackage{multirow}
\newcommand{\eg}{\textit{e.g.}}
\newcolumntype{C}[1]{>{\centering\let\newline\\\arraybackslash\hspace{0pt}}m{#1}}
\usepackage{algorithm} 
\usepackage{algpseudocode}
\usepackage{wrapfig}
\usepackage{caption}
\usepackage{textcomp}
\usepackage{booktabs}       
\usepackage{amsfonts}       
\usepackage{nicefrac}       
\usepackage{xcolor}         

\newcommand{\jyp}[1]{{\color[rgb]{0,0,0}{#1}}}

\title{Generative Subgraph Retrieval for \\Knowledge Graph\textendash{Grounded} Dialog Generation}

\author{Jinyoung Park$^1$\thanks{Part of this work was done during an internship at Amazon AGI.} \quad Minseok Joo$^1$ \quad Joo-Kyung Kim$^2$\thanks{Co-corresponding authors.} \quad Hyunwoo J. Kim$^{1\dagger}$ \\
        $^1$Korea University,  $^2$Amazon AGI\\
        {\tt \{lpmn678, wlgkcjf87, hyunwoojkim\}@korea.ac.kr} \quad {\tt jookyk@amazon.com}\\}

\begin{document}
\input{notation}
\maketitle
\begin{abstract}

\input{0_Abstract/0_abstract}
\end{abstract}

\section{Introduction}
\input{1_Introduction/1_0_introduction}

\section{Related Works}
\input{2_RelatedWorks/2_0_related_works}

\section{Methods}
\input{3_Methods/3_0_methods}
\subsection{KG\textendash{Grounded} Dialog Generation}
\label{sec:3.1}
\input{3_Methods/3_1_preliminaries}
\subsection{Generative Subgraph Retrieval}
\input{3_Methods/3_2_multihop_knowledge_paths}
\subsection{Response Generation}
\input{3_Methods/3_4_arch}
\subsection{Training DialogGSR}
\label{sec:3.4}
\label{sec:method_training}
\input{3_Methods/3_5_training}

\section{Experiments}
\label{sec:exp}
\input{4_Experiments/4_0_experiments}
\subsection{Experimental Setup}

\input{4_Experiments/4_2_setup}
\subsection{Experimental Results}
\input{4_Experiments/4_3_results}

\subsection{Analysis}
\label{sec:analysis}
\input{5_Analysis/5_0_analysis}

\section{Conclusion}
\input{6_Conclusion/6_0_conclusion}

\section*{Acknowledgements}
This work~(for authors: HJK, JP, and MJ) was partly supported by ICT Creative Consilience Program through the Institute of Information \& Communications Technology Planning \& Evaluation~(IITP) grant funded by the Korea government~(MSIT)~(IITP-2024-RS-2020-II201819, 10 \%), and National Research Foundation of Korea grant (NRF-2023R1A2C2005373, 90 \%).

\bibliography{main_cam}

\appendix
\label{sec:appendix}
\input{appendix}
\end{document}

%% file: notation.tex
\newcommand{\DialHist}{\mathcal{D}} 
\newcommand{\Word}{w}
\newcommand{\Token}{x}
\newcommand{\hx}{h_{\TokenSeq}}
\newcommand{\TokenSeq}{\boldsymbol{x}}
\newcommand{\kpTokenSeq}{\boldsymbol{z}}
\newcommand{\enc}{\text{enc}}
\newcommand{\hb}{\mathbf{h}}
\newcommand{\xb}{\boldsymbol{x}}
\newcommand{\zb}{\boldsymbol{z}}
\newcommand{\MLPx}{\mathbf{MLP}_{\boldsymbol{x}}}
\newcommand{\MLPp}{\mathbf{MLP}_{p}}
\newcommand{\Voca}{\mathcal{V}}
\newcommand{\Mask}{\texttt{[MASK]}}
\newcommand{\Sep}{\texttt{[SEP]}}
\newcommand{\Special}{\texttt{[SP\textsubscript{\texttt{t}}]}}
\newcommand{\Specialone}{\texttt{[SP\textsubscript{\texttt{1}}]}}
\newcommand{\Specialtwo}{\texttt{[SP\textsubscript{\texttt{2}}]}}
\newcommand{\Specialthree}{\texttt{[SP\textsubscript{\texttt{3}}]}}
\newcommand{\Specialfour}{\texttt{[SP\textsubscript{\texttt{4}}]}}
\newcommand{\Specialfive}{\texttt{[SP\textsubscript{\texttt{5}}]}}
\newcommand{\Specialsix}{\texttt{[SP\textsubscript{\texttt{6}}]}}
\newcommand{\Speciallast}{\texttt{[SP\textsubscript{\texttt{l+1}}]}}
\newcommand{\KG}{\mathcal{G}}
\newcommand{\EntSet}{\mathcal{E}}
\newcommand{\RelSet}{\mathcal{R}}
\newcommand{\PathSet}{\mathcal{P}}
\newcommand{\Ent}{e}
\newcommand{\Head}{h}
\newcommand{\Tail}{t}
\newcommand{\Rel}{r}
\newcommand{\OutSeq}{\boldsymbol{y}}
\newcommand{\Out}{y}
\newcommand{\Path}{p}
\newcommand{\TripleSet}{\mathcal{T}}
\newcommand{\PathPrompt}{\Psi}
\newcommand{\ie}{\textit{i.e.}}
\newcommand{\MentionSet}{\mathcal{M}}


    

%% file: 0_Abstract/0_abstract.tex
Knowledge graph{\textendash}grounded dialog generation requires retrieving a dialog-relevant subgraph from the given knowledge base graph and integrating it with the dialog history.
Previous works typically represent the graph using an external encoder, such as graph neural networks, and retrieve relevant triplets based on the similarity between single-vector representations of triplets and the dialog history. 
However, these external encoders fail to leverage the rich knowledge of pretrained language models, and the retrieval process is also suboptimal due to the information bottleneck caused by the single-vector abstraction of the dialog history.
In this work, we propose Dialog generation with Generative Subgraph Retrieval (DialogGSR), which retrieves relevant knowledge subgraphs by directly generating their token sequences on top of language models. 
For effective generative subgraph retrieval, we introduce two key methods: (i) structure-aware knowledge graph linearization with self-supervised graph-specific tokens and (ii) graph-constrained decoding utilizing graph structural proximity-based entity informativeness scores for valid and relevant generative retrieval. 
DialogGSR achieves state-of-the-art performance in knowledge graph{\textendash}grounded dialog generation, as demonstrated on OpenDialKG and KOMODIS datasets.


%% file: 1_Introduction/1_0_introduction.tex
The goal of dialog generation is to generate an informative and appropriate response given an input dialog.
Pretrained Language Models~(PLMs) have demonstrated promising performance on the dialog generation~\cite{roberts2019exploring,touvron2023llama,achiam2023gpt}.
However, they often generate irrelevant, factually incorrect, or hallucinatory responses since the generation process heavily depends on the internal parameters of the language models~\cite{lewis2020retrieval,shuster2021retrieval}.
To mitigate these issues, several studies~\cite{wang2020improving,zhao2020low} have explored knowledge-grounded dialog generation models, which incorporate external knowledge to generate more factually accurate responses.
Some approaches utilize unstructured texts such as Wikipedia articles~\cite{dinan2019wizard} and internet web pages~\cite{ghazvininejad2018knowledge} while others~\cite{moon2019opendialkg,galetzka2021space,tuan2022towards,kang2022knowledge} leverage structured knowledge graphs~(KGs) to capture both the relational and semantic information for grounding dialog responses.

{
Many existing knowledge graph{\textendash}grounded dialog generation models~\cite{tuan2022towards,kang2022knowledge} employ encoder-based retrieval methods.
They encode the dialog history into a single vector and then use it on another encoder~(\eg,~bi-encoder) to retrieve relevant triplets from the KG.
However, this approach can lead to an information bottleneck due to the limited capacity of a single vector to represent long and complex multi-turn dialogs~\cite{humeau2020poly,de2021autoregressive,lee2022generative}.
Moreover, these methods~\cite{galetzka2021space,tuan2022towards,kang2022knowledge} often rely on separate models, such as graph neural networks (GNNs), to encode the knowledge graphs, which limits the integration of natural language comprehension capabilities of PLMs.
}

Recent studies~\cite{lee2022generative,sun2023generative} have addressed the information bottleneck issue by applying generative retrieval methods, which cast retrieval as an autoregressive generation process to facilitate direct interactions between query context and knowledge paragraphs.
Despite this progress, most generative retrieval works focus solely on natural language-based knowledge, employing conventional token representations and decoding strategies, which do not fully capture the structure and properties of knowledge graphs.

{To address the aforementioned issues, we propose \textbf{Dialog} Generation model with \textbf{G}enerative \textbf{S}ubgraph \textbf{R}etrieval~(\textbf{DialogGSR}), which integrates generative subgraph retrieval with response generation. 
Our proposed method adopts two key graph-specialized techniques: (1) a structure-aware knowledge graph linearization for effective graph representation and (2) graph-constrained decoding for valid subgraph retrieval.}
\textcolor{black}{Our knowledge graph linearization approach introduces a small set of special token embeddings to account for both the structural positioning of knowledge entities and the reverse relationships between them. By self-supervising these special tokens using a knowledge graph reconstruction loss, the method effectively represents the knowledge graph.}
The graph-constrained decoding facilitates autoregressively retrieving the knowledge considering the graph structural information, thus generating valid and relevant knowledge subgraphs.
Since DialogGSR utilizes pretrained language models for both subgraph retrieval and dialog generation, it leverages the pretrained language models' internal knowledge in both tasks.

We evaluate DialogGSR on two KG{\textendash}grounded dialog generation datasets: OpenDialKG~\cite{moon2019opendialkg} and KOMODIS~\cite{galetzka2020corpus}.
Our proposed method shows the best performance on both benchmark datasets.

Our contributions are three-fold as follows:
\begin{itemize}
    \item We propose Dialog generation with Generative Subgraph Retrieval~(DialogGSR), which retrieves the relevant knowledge subgraphs by generating their token sequences. 
    \item We design knowledge graph linearization for effective graph representations and graph-constrained decoding for retrieving valid and relevant subgraphs.
    \item We show the state-of-the-art response generation performance on two benchmark datasets, OpenDialKG and KOMODIS.
\end{itemize}

%% file: 2_RelatedWorks/2_0_related_works.tex
\input{Figure_tex/temp.tex}

\subsection{Generative Retrieval}

Retrieving relevant information from a large corpus such as a text corpus or a knowledge base is crucial in many tasks~\cite{ChenFWB17,ThorneVCM18,lewis2020retrieval,IzacardG21}.
Recent studies~\cite{de2021autoregressive,bevilacqua2022autoregressive,wang2022neural,lee2022generative,lee2023contextualized} have demonstrated that generative retrieval models can be more effective than conventional encoder-based retrieval models. 
They cast retrieval tasks as generation tasks, where relevant sequences are generated rather than retrieved given input queries.
Several studies~\cite{chen2022gere,thorne2022data,lee2022generative,yu2023generate,xu-etal-2023-unsupervised,luo2024reasoning} have shown the effectiveness of generative retrieval in various knowledge-intensive natural language processing tasks.
Motivated by these works, we propose a generative subgraph retrieval model with knowledge graph linearization and graph-constrained decoding for effective graph representation and generation.

\subsection{Knowledge-Grounded Dialog Generation}
Many language generation approaches leverage pretrained language models (PLMs)~\cite{radford2019language,devlin2018bert,roberts2019exploring,thoppilan2022lamda,touvron2023llama,achiam2023gpt}, showing strong performance.
However, they often suffer from the hallucination issue~\cite{duvsek2018findings,balakrishnan2019constrained,duvsek2020evaluating}, which generates plausible but factually wrong responses since they rely on the models' internal parameters.
To address this problem, recent works~\cite{moon2019opendialkg,dinan2019wizard,lian2019learning,park2023relation} have proposed to augment the models with external knowledge sources.
This approach is effective in generating factually accurate responses in various language generation tasks~\cite{fernandes2018structured,huang2020knowledge,yasunaga2021qa,yu2022jaket,zhang2022greaselm}.
Regarding dialog generation, various works incorporate external knowledge graph into the generation~\cite{moon2019opendialkg,zhou2018commonsense,tuan2019dykgchat,zhang2019grounded,zhou2021earl}. 
For instance, Space Efficient~\cite{galetzka2021space} proposes an efficient method to encode knowledge triplets.
RHO~\cite{ji2022rho} generates responses with the dialog history and knowledge graph represented by graph embedding methods~(\eg, TransE \cite{NIPS2013_1cecc7a7}).
DiffKG~\cite{tuan2022towards} uses a graph reasoning encoder on top of sparse matrices for graph representations.
SURGE~\cite{kang2022knowledge} applies GNNs to retrieve context-relevant subgraphs. 
Different from these works, our work autoregressively retrieves the context-relevant subgraphs and then generates knowledge-grounded dialogs without separate knowledge graph modules.

%% file: Figure_tex/temp.tex
\begin{figure*}[t]
\centering
\includegraphics[width=1.0\textwidth]{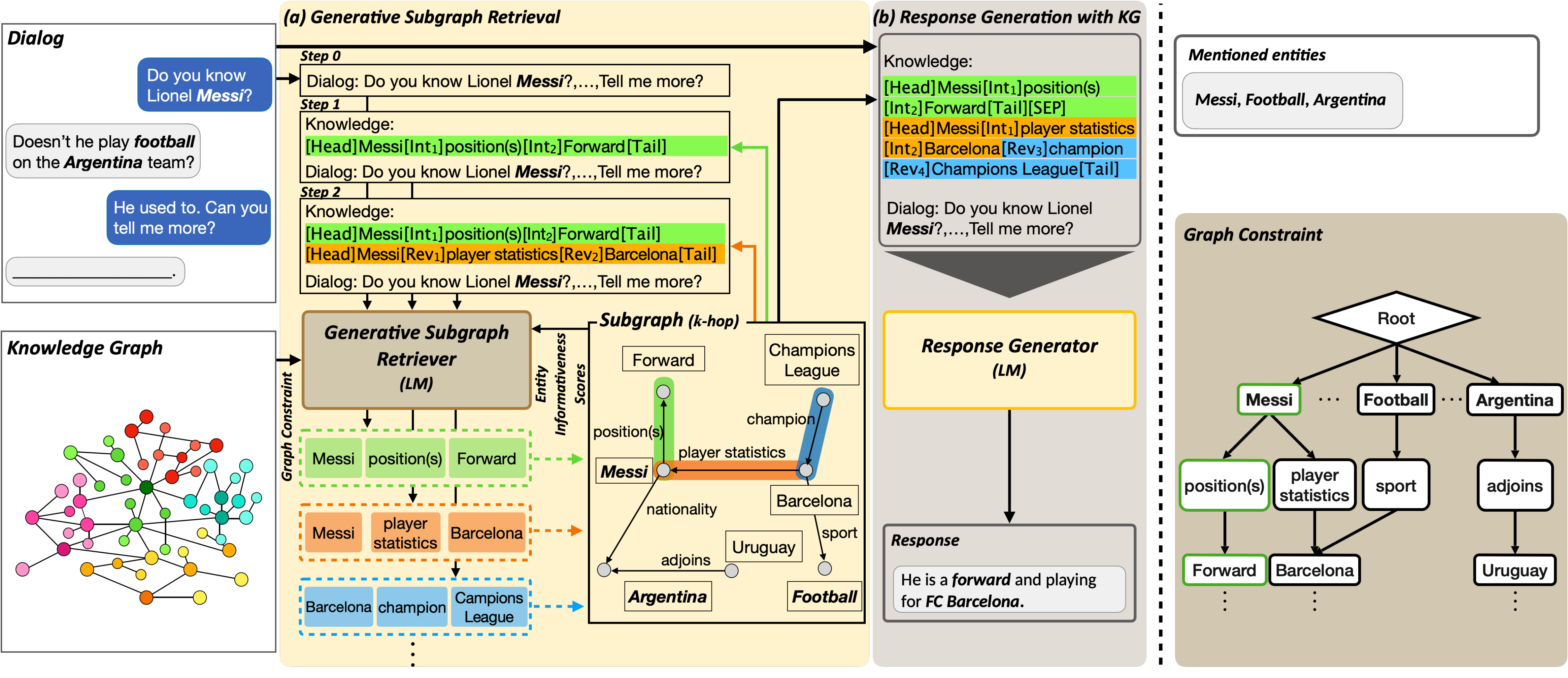}
\caption{
The overall inference process of DialogGSR.
DialogGSR consists of a generative subgraph retriever and response generator.
(a) Generative subgraph retrieval autoregressively retrieves subgraphs via generative subgraph retriever with graph-constrained decoding based on entity informativeness score.
In step 0, given the dialog, GSR retrieves the most relevant triplets by referring to the graph constraint.
In step 1, given the dialog and the prompt-augmented triplet, we generatively retrieve the next triplets.
(b) Resposne generator generates the responses with the dialog and the prompt-augmented generated subgraph.
}
\label{fig:main}
\end{figure*}


%% file: 3_Methods/3_0_methods.tex
We propose a retrieval-augmented dialog generation approach that retrieves contextually relevant subgraphs from knowledge graphs to generate better responses.
Our model, \textbf{Dialog} Generation model with \textbf{G}enerative \textbf{S}ubgraph \textbf{R}etrieval~(DialogGSR) consists of a generative subgraph retriever and a response generator.
We first define the task of knowledge graph{\textendash}grounded dialog generation~(Sec.~\ref{sec:3.1}).
Next, we propose \textbf{G}enerative \textbf{S}ubgraph \textbf{R}etrieval~(GSR), which autoregressively retrieves subgraph by applying structure-aware knowledge graph linearization and graph-constrained decdoing~(Sec.~\ref{sec:3.2}).
We then present a response generator, which performs subgraph{\textendash}grounded dialog generation~(Sec~\ref{sec:3.3}).
Finally, we provide the training details of DialogGSR including our self-supervised knowledge graph reconstruction loss~(Sec~\ref{sec:3.4}).
The inference process of DialogGSR is illustrated in Figure~\ref{fig:main}.

%% file: 3_Methods/3_1_preliminaries.tex
The goal of knowledge graph{\textendash}grounded dialog generation is to generate a dialog response by jointly reasoning over a dialog history and a knowledge graph.
We represent a dialog history as a token sequence, $\TokenSeq = \left[\Token_1, \Token_2, \dots, \Token_{n} \right]$, where $\Token_i \in \Voca$ is the $i$-th token of the dialog history and $\Voca$ denotes the vocabulary set.
A knowledge graph is defined as $\KG = \left(\EntSet, \RelSet, \TripleSet \right)$, where $\EntSet$ is the set of entities and $\RelSet$ is the set of relations.
$\TripleSet$ denotes the set of triplets, $(\Ent_h, \Rel, \Ent_t) \in \TripleSet$, each of which are composed of a head entity $\Ent_h \in \EntSet$, a tail entity $\Ent_t \in \EntSet$, and a relation $\Rel \in \RelSet$ between the two entities.
\jyp{We use $k$-hop subgraph linked to the entities mentioned in the input dialog as retrieval candidates following previous works~\cite{kang2022knowledge}.}
The example of a extracted candidate subgraph is in Figure~\ref{fig:subgraph}.
We formulate knowledge graph{\textendash}grounded dialog generation as follows:
\begin{equation}
\label{eq:problem}
    p_{\theta}(\OutSeq |\TokenSeq, \KG) = \prod_{j=1}^t p_{\theta}(\Out_j | \TokenSeq, \OutSeq_{<j}, \KG),
\end{equation}
where $\OutSeq=\left[y_1, y_2, \dots, y_t \right]$ is the output response, $t$ is the length of the response, and $\OutSeq_{<j}=\left[\Out_1, \dots \Out_{j-1} \right]$ denotes the generated sequence at the previous time steps. 
Since a KG can include a huge number of irrelevant entities and relations, KG-grounded dialog generation works generally retrieve subgraphs related to the dialog context for the efficiency and effectiveness.

%% file: 3_Methods/3_2_multihop_knowledge_paths.tex
\label{sec:3.2}
We introduce \textbf{G}enerative \textbf{S}ubgraph \textbf{R}etrieval (GSR), which autoregressively retrieves a knowledge subgraph $\hat{\KG}$.
Since a knowledge subgraph can be represented as a set of triplets, retrieving sequences of knowledge triplets is equivalent to subgraph retrieval.
Many subgraph retrieval methods in dialog generation~\cite{zhang2022subgraph,kang2022knowledge} compute the relevance score between the dialog history and each knowledge triplet and retrieve the triplets with the highest scores.

However, these methods often suffer from the information bottleneck problem~\cite{izacard2020memory, luan2021sparse}, as they encode long, multi-turn dialog histories into a single fixed-length vector, which has a limited capacity to accurately represent complex multi-turn dialogs.
Moreover, these approaches require independent knowledge graph encoders to represent knowledge graphs, which cannot fully leverage the pre-trained knowledge embedded in the pretrained language models.

To address these limitations, generative subgraph retrieval casts graph retrieval as a graph generation, enabling more direct interaction between the dialog context and the knowledge graph by representing the graph with a token sequence.
For effective generative retrieval, our GSR model incorporates two novel techniques: (1) Structure-aware knowledge graph linearization, which converts the knowledge graph into token sequences enriched with learnable special tokens that capture the connectivity and reverse relations between entities, and (2) Graph-constrained decoding, which ensures the language model to generate valid knowledge subgraphs by predicting the next tokens based not only on the language model's scores but also on the relational proximities of entities within the graph.

\paragraph{Structure-aware knowledge graph linearization.}
The goal of knowledge graph linearization is to convert a knowledge graph into a token sequence comprehensible to language models.
Our structure-aware knowledge graph linearization augments a sequence of knowledge graph tokens with graph-specific learnable special tokens to help the language model understand the graph's structural information without separate graph encoders.
Different from prior graph linearization methods such as \citet{xu-etal-2023-unsupervised}, which do not take into account multi-hop graph connections and reverse relations, our structure-aware knowledge graph linearization better captures and effectively represents the underlying structures of knowledge graphs.

Specifically, if there are connected triplets~(\eg, $(e_1, r_1, e_2)$ and $(e_2, r_2, e_3)$), we efficiently represent the path as $\texttt{[Head]}$ $\Ent_1$ $\texttt{[Int{\textsubscript{\texttt{1}}}]}$ $\Rel_1$ $\texttt{[Int{\textsubscript{\texttt{2}}}]}$ $\Ent_2$ $\texttt{[Int{\textsubscript{\texttt{3}}}]}$ $\Rel_2$ $\dots\Ent_{l+1}$ $\texttt{[Tail]}$.
To represent multiple disconnected triplets or paths, we insert $\texttt{[SEP]}$ between them.
For more expressive representations of the special tokens, we use multiple consecutive tokens to represent each of $\texttt{[Int],[Rev]}$, which improves the performance as in Section \ref{sup_sec:aqa}.

Additionally, since a knowledge graph can contain reverse relations, representing them is crucial in knowledge graph processing~\cite{feng2020scalable,qi2023investigation,zhu2024dyval}.
Therefore, we introduce another special token $\texttt{[Rev]}$ for reverse relations when (1) there is a mentioned entity that is the tail of a triplet because the decoding always starts with one of the mentioned entities, or (2) two triplets are connected with opposite directions (\eg, $(e_1, r_1, e_2)$ and $(e_3, r_2, e_2)$). 
We effectively represent reverse relations by adding special tokens $\texttt{[Rev{\textsubscript{\texttt{1}}}]}$ and $\texttt{[Rev{\textsubscript{\texttt{2}}}]}$ without modifying the relation tokens. For example, given a triplet $(e_3, r_2, e_2)$, the corresponding triplet with the reverse relation $(e_2, \tilde r_2, e_3)$ is represented as $\texttt{[Head]} \ \Ent_2 \ \texttt{[Rev{\textsubscript{\texttt{1}}}]} \ \Rel_2 \ \texttt{[Rev{\textsubscript{\texttt{2}}}]}  \ \Ent_{3} \ \texttt{[Tail]}$.

In sum, we represent the subgraph $\hat{\KG}$ as the concatenation of the knowledge paths converted with the special tokens as follows:
\begin{equation}
\begin{split}
\kpTokenSeq_{\hat{\KG}} =  &\texttt{[Head]}\Ent_1\texttt{[Int{\textsubscript{\texttt{1}}}]}\Rel_1\dots \\ &\Ent_{l+1}\texttt{[Tail]} \Sep\texttt{[Head]}\Ent_k\cdots.
\end{split}
\label{eq:kpg}
\end{equation}
All the special tokens are learnable with soft prompting.
\textcolor{black}{They are learned with both downstream task loss and knowledge graph reconstruction loss, which will be introduced in Section~\ref{sec:method_training}.
Our structure-aware knowledge graph linearization with the special tokens helps the language model capture knowledge graph information without any separate knowledge graph encoders, which leads to the full utilization of the power of PLMs.
}

\paragraph{Graph-constrained decoding.}
The language model is prone to generating invalid or irrelevant subgraphs due to its bias, often disregarding the knowledge graph structures~\cite{de2021autoregressive,chen2022relation}.
To address this issue, we introduce a graph-constrained decoding method that ensures the generation of valid and relevant subgraphs.
Formally, given a dialog $\boldsymbol{x}$ and the previously generated segments of linearized knowledge path ${\pi}_{<t}$, the log probability of the next token $w$ is computed with $\log p_{\text{vocab}}\left(w|\boldsymbol{x}, {\pi}_{<t}, C_{\mathcal{M}} \right)$.
Here, $C_{\mathcal{M}}$ represents a prefix tree derived from the ego-graph~\cite{zhu2021transfer} of a set of mentioned entities $e_m \in \mathcal{M}$ as depicted in Figure~\ref{fig:main}~(right).
\jyp{
The mentioned entities are the entities that appear in the input dialog history and correspond to entities in the knowledge graph~\cite{kang2022knowledge}.
For example, given the dialog ``Do you know Lionel Messi?" in Figure~\ref{fig:main}, the entity `Messi' is a mentioned entity since it exists in the knowledge graph.
}
The next token prediction probability $p_{\text{vocab}}$ is restricted to tokens within the valid set defined by the constraint $C_{\mathcal{M}}$~(\ie, $({\pi}_{<t}, w) \in \mathcal{C}_{\mathcal{M}}$).
This constraint ensures that only valid knowledge subgraphs are generated.

In addition, to account for the importance of each entity in the knowledge graph, we introduce a graph-based next-token prediction probability, which is defined as:
\begin{equation}
\label{eq:constraint}
\begin{split}
&\log\tilde{p}(w|\boldsymbol{x}, {\pi}_{<t}, C_{\mathcal{M}}) \\
&= \alpha \cdot \log p_{\text{vocab}}\left(w|\boldsymbol{x}, {\pi}_{<t}, C_{\mathcal{M}} \right)\\&+ {(1-\alpha)}\cdot \log p_{\text{graph}}\left(w| {\pi}_{<t}, C_{\mathcal{M}} \right),
\end{split}
\end{equation}
where $p_{\text{graph}}$ is the probability of predicting the next token based on graph structure, and $\alpha$ is a hyperparameter controlling the balance between the language model and graph-based predictions. 
If the next token $w$ corresponds to a tokenized entity, the probability $p_{\text{graph}}$ is defined as:
\begin{equation}
    p_{\text{graph}}\left(w| {\pi}_{<t},  C_{\mathcal{M}}\right) \propto \mathcal{S}(e_i, \mathcal{M}),
\end{equation}
where $\mathcal{S}\left(e_i, \mathcal{M} \right)$ is the entity informativeness score of entity $e_i$ with respect to the mentioned entity set $\mathcal{M}$. 
 In cases where all entities have identical informativeness scores, the next token prediction is only dependent on $p_{\text{vocab}}$.

To capture the structural proximity between entity $e_i$ and mentioned entities $e_m \in \mathcal{M}$ on the graph, we define the structure-based entity Informative Score~(IS) as
\begin{equation}
    \mathcal{S}(e_i, \mathcal{M}) = \frac{1}{\lvert \mathcal{M}\rvert} \sum_{e_m\in \mathcal{M}}s(e_i, e_m),
\end{equation}
where $s(e_i, e_m)$ denotes the graph structural proximity between entity $e_i$ and $e_m$.
The proximity can be measured using methods such as the shortest path and common neighbors~\cite{katz1953new,brin1998pagerank,gasteiger2018predict}.
A typical approach for measuring graph structural proximity is counting the number of connections between node pairs, which can be defined as $s_{\text{con}}(e_i, e_m) = \sum_{\mathcal{N}(e_m)} \mathbf{1}(e_i=e_m)$, where $\mathcal{N}(e)$ is the neighborhood set of entity $e$.

However, the connection-based proximity measurement fails to account for multi-hop relations.
To address this, we introduce a Katz index{\textendash}based entity informativeness score~($\mathcal{IS}_{\text{katz}}$)~\cite{katz1953new}, formulated as follows:
\begin{equation}
\label{eq:katz}
\mathcal{IS}_{\text{katz}}(e_i, \mathcal{M}) = \frac{1}{\lvert \mathcal{M}\rvert} \sum_{e_m\in \mathcal{M}}\sum_{k=1}^K \beta^k(\mathbf{A}^k)_{i,m},
\end{equation}
where $\mathbf{A}$ is the adjacency matrix of graph $\mathcal{G}$, $K$ denotes the maximum length of knowledge paths and $\beta^k$ means a weight of knowledge path of length $k$.
Since the term $\mathbf{A}^k$ represents the number of paths between entity $e_i$ and $e_m$, this Katz index{\textendash}based entity informativeness score enables multi-hop relationship modeling, in contrast to the simple connection-based metrics.

%% file: 3_Methods/3_4_arch.tex
\label{sec:3.3}
\jyp{After retrieving the subgraphs, we generate a response based on both the dialog history and the retrieved subgraphs.}
To incorporate the retrieved knowledge subgraph $\hat{\mathcal{G}}$, we first apply the knowledge graph linearization to convert $\hat{\mathcal{G}}$ into a sequence of tokens, $\boldsymbol{z}_{\hat{\KG}}$.
This linearized subgraph is then concatenated with the dialog history $\boldsymbol{x}$, forming the input sequence for the dialog generation model as
\begin{equation}
    \hat{\boldsymbol{x}} = \left[\boldsymbol{z}_{\hat{\KG}};\boldsymbol{x} \right],
\end{equation}
where $[;]$ denotes concatenation operation.
The combined sequence is fed into the response generation model to get the final response $\boldsymbol{y}$.
By augmenting the dialog input with the knowledge graph, this method ensures that the generated response is both contextually relevant and knowledge-grounded.

%% file: 3_Methods/3_5_training.tex
\label{sec:3.5}

Our DialogGSR is trained in a multi-stage process.
The training process consists of: (1) self-supervision through knowledge graph reconstruction, (2) training the generative subgraph retriever, and (3) optimizing the response generation model. 
These stages work in synergy to ensure the model effectively retrieves knowledge from graphs and generates coherent, knowledge-grounded responses.
We also train the response generator by minimizing response generation loss.

\paragraph{Knowledge graph reconstruction.}
Inspired by masked language modeling techniques~\cite{roberts2019exploring,devlin2018bert}, we propose a self-supervised learning approach to learn the special tokens by masking either an entity token or a relation token in the token sequence of each knowledge path and reconstructing it.
Specifically, we first sample $k$-hop path $\KG^\prime$ from the knowledge source graph $\KG$ and convert it into token sequence~$\boldsymbol{z}_{\KG^\prime}$.
During training, we randomly mask out either an entity token or a relation token from the sequence. 
The loss is formulated as
\begin{equation}
    \mathcal{L}_{\text{GraphRecon}} = -\log p(\zb_{\KG^\prime} | \hat{\zb}_{\KG^\prime}),
\end{equation}
where $\boldsymbol{z}_{\KG^\prime}$ is the token sequence of a sampled path and $\hat{\boldsymbol{z}}_{\KG^\prime}$ is its randomly masked sequence.
For example, a knowledge triplet $\zb_p =\langle$ `Scarlet Letter', `written by', `N.Hawthorne' $\rangle$ can be randomly masked as 
\begin{equation*}
\begin{split}
&\langle \texttt{<M>, `written by', `N.Hawthorne'} \rangle \\
&\langle \texttt{`Scarlet Letter', <M>, `N.Hawthorne'} \rangle\\
&\langle \texttt{`Scarlet Letter', `written by', <M>} \rangle.
\end{split}
\end{equation*}
Note that masking is done at the entity or relation level as done in \citet{roberts2019exploring}. 
By minimizing the graph reconstruction loss, our framework self-supervise the special tokens $\texttt{[Head],[Int],[Rev],[Tail]}$ in \eqref{eq:kpg}, resulting in better knowledge graph representations.
All the other parameters are frozen during this stage.

\paragraph{Knowledge subgraph retrieval.}
We train our generative subgraph retriever (GSR) to identify relevant subgraphs for dialog generation.
Unlike conventional retrieval methods, our approach frames retrieval as a generation task, enabling a more seamless integration with the dialogue context. The loss is defined as follows:
\begin{equation}
    \mathcal{L}_{\text{Ret}} = \mathbb{E}_{\TokenSeq} \left [ -\log p\left({\KG}^\star | \TokenSeq \right) \right ] 
\end{equation}
where $\KG^{\star}$ is the gold subgraph and $\TokenSeq$ is the dialog context
We use cross-entropy loss to train the retriever, ensuring it generates subgraphs that are both relevant and informative.
\paragraph{Response generation.}
The final stage of training DialogGSR is response generation.
We generate dialog responses with dialog history $\TokenSeq$ and context-relevant knowledge subgraphs $\hat\KG$ retrieved from GSR. 
The response generation loss is defined as follows:
\begin{equation}
\mathcal{L}_{\text{Gen}} = \mathbb{E}_{\xb} \left [ -\log p\left(\OutSeq^\star | \TokenSeq, \hat{\KG} \right) \right ],
\end{equation}
where $\OutSeq^\star$ is the golden response.

%% file: 4_Experiments/4_0_experiments.tex
\input{Tables/main_table}
\input{Tables/komodis_table}

\input{Tables/retrieval_table}

\input{Tables/supp_human_eval}
\input{Tables/ablation}
\input{Tables/llama_table}

In this section, we evaluate the effectiveness of the proposed DialogGSR on knowledge graph{\textendash}grounded dialog generation.
We first introduce the two datasets~(OpenDialKG~\cite{moon2019opendialkg} and KOMODIS~\cite{galetzka2020corpus}), and the experimental setup and metrics.
Then, we demonstrate the effectiveness of DialogGSR on the two benchmark datasets.
Lastly, we provide ablation studies, and analyses of our DialogGSR. 

\subsection{Datasets}
\textbf{OpenDialKG} is an open-domain dialog dataset, which consists of 15K dialogs with 91K turns and 1.12M triplets from Freebase knowledge graph~\cite{bast2014easy}.
The knowledge graph has 1,190,658 triplets, 100,813 entities, and 1,358 relations.
There are 49\% of the turns having gold knowledge triplets.
Following \cite{galetzka2020corpus}, we randomly split the samples into train~(70\%), validation~(15\%), and test~(15\%) sets.
We evaluate the response generation and retrieval performance of our DialogGSR with other baselines using OpenDialKG dataset.

\noindent\textbf{KOMODIS} is a closed-domain dialog dataset that consists of 7.5k dialogs with 103k turns and the corresponding KG, which contains 88K triplets.
Following \cite{moon2019opendialkg,kang2022knowledge,galetzka2020corpus}, we randomly split the dialogs into train~(70\%), validation~(15\%), and test~(15\%) sets for KOMODIS dataset, too.
With KOMODIS dataset, we evaluate the response generation performance of our DialogGSR with other baselines following \cite{kang2022knowledge,galetzka2021space}.

%% file: Tables/main_table.tex
\begin{table*}[ht!]
    \centering
    \small
    \begin{tabular}{l|cccc|ccc|c|cc}
        \toprule
        \multirow{2}{*}{Method}  &  \multicolumn{4}{c|}{\textbf{BLEU}} & \multicolumn{3}{c|}{\textbf{ROUGE}}& \textbf{Unigram} &\multicolumn{2}{c}{\textbf{KQA}}\\
          &  B-1 & B-2 & B-3 & B-4 & R-1 & R-2 &R-L&  F1& EM& F1\\
         \midrule 
         \midrule
         T5~(w/o KG)  &  15.79 & 9.19 & 5.61 & 3.43 & 19.67& 7.13& 19.02& 22.21 & 12.25& 20.69\\
         Space Efficient~(series) &16.15&10.03&6.66&4.50&21.15 &8.56&20.44& 24.55& 36.60& 42.64\\
         Space Efficient~(parallel)  &16.33& 10.22& 6.81& 4.64& 21.42& 8.85& 20.68 & 24.87& 38.54 & 44.34\\
         EARL  &11.49 & 6.34 & 4.06 & 2.75 & 15.36 & 4.37 & 14.61 & 16.88 & 32.47 & 35.88\\
         DiffKG  &15.68 & 9.13 & 5.60 & 3.46 & 19.50 & 7.07 & 18.84 & 22.26 & 12.25 & 20.99\\
         SURGE~(unsup.)  &17.77 &11.30 &7.69 &5.36  & 21.64 & 9.14 & 20.75 & 25.24 & 48.49 & 55.77\\
         SURGE~(semi-sup.) &17.70 & 11.21 & 7.61 & 5.28 &21.43 & 8.85 & 20.57& 25.07 & 51.00 & 57.63\\
         SURGE~(contrastive) &17.29 & 11.04 & 7.54 & 5.28 &21.35 & 8.98 & 20.48& 25.10& 50.45 & 57.70\\
         \midrule
         \textbf{DialogGSR~(Ours)} &\textbf{19.30}&\textbf{12.10}& \textbf{8.30}& \textbf{5.83} &\textbf{22.32} & \textbf{9.24} & \textbf{21.23} & \textbf{25.50}& \textbf{54.61}& \textbf{60.57}\\
        \bottomrule
    \end{tabular}
    \vspace{-5pt}
    \caption{Response generation performance comparison on OpenDialKG dataset. 
    }
\label{tab:opendialkg}
\end{table*}

%% file: Tables/komodis_table.tex
\begin{table}
\centering
    \small
    \begin{tabular}{l|c|c|c}
        \toprule
        {Method} &  BLEU & ROUGE & F1\\
         \midrule 
         \midrule
         T5~(w/o KG) & 7.58 & 18.54 & 16.60\\
         Space Efficient~(series) & 8.34 & 22.36 & 17.37\\
         Space Efficient~(parallel) & 9.33 & 22.80 & 17.72\\
         SURGE~(unsup.) & 11.46 & 23.49 & 18.70\\
         SURGE~(semi-sup.) & 11.28 & 23.58 & 18.68\\
         SURGE~(contrastive)  & 11.51 & 24.13 & 19.51\\
         \midrule
         \textbf{DialogGSR~(Ours)} & \textbf{11.96} & \textbf{24.47} &\textbf{19.60}\\
        \bottomrule
    \end{tabular}
    \vspace{-5pt}
    \caption{Experimental results on KOMODIS dataset.
}
\label{tab:komodis}
\end{table}

%% file: Tables/retrieval_table.tex
\begin{table}
\centering
\small
\begin{tabular}{l|cc}
\toprule
    Method& path@1&path@3\\
     \midrule
     \midrule
     Seq2Seq   & 3.1&18.3\\
     Tri-LSTM  & 3.2 & 14.2 \\
     EXT-ED  & 1.9 & 5.8 \\
     DialKG Walker & 13.2 & 26.1 \\
     AttnFlow & 17.37&24.84 \\
     AttnIO & 23.72& 37.53\\
     DiffKG & 26.12 &44.50 \\
     SURGE &16.76 &28.64 \\
     \midrule
     DialogGSR~(Ours)  & \textbf{28.96}&\textbf{46.76}\\
     \bottomrule
    \end{tabular}
    \vspace{-5pt}
\caption{\label{tab:retrieval}Retrieval performance on OpenDialKG.}
\end{table}

%% file: Tables/supp_human_eval.tex
\begin{table}[t]
    \small
    \centering
    {
    
    \begin{tabular}{c|cc}
        \toprule
          \textbf{Method} & \textbf{DialogGSR~(Ours)} & SURGE\\
         \midrule 
         \midrule
                       
        Consistency&\textbf{2.57 (0.168)} &2.41 (0.196)        \\
        Informativeness&\textbf{2.28 (0.136)} & 1.81 (0.260) \\
        Fluency &\textbf{2.64 (0.200)} & 2.53 (0.286)  \\
        \bottomrule
    \end{tabular}
    }
    \caption{Human evaluation results. () indicates standard deviation.}
    \label{tab:humaneval}
\end{table}

%% file: Tables/ablation.tex
\begin{table*}[t]
    \small
    \centering
    \begin{tabular}{l|l|cccc|cc}
        \toprule
          {Graph Const.} & {Special tokens} &  {B-1} & {B-2} & {B-3} & {B-4} & path@3\\
         \midrule 
        w/o Const.  & with Special tokens (w/o Recon.) & 17.02&10.96& 7.53& 5.25  &10.00\\
        Hard Const. & w/o Special tokens &  18.44 & 11.68 & 7.93 & 5.44  & 35.83\\
        Hard Const. & with Special tokens (w/o Recon.) &  18.77 & 11.74 & 8.03 & 5.48  & 39.53\\
        Hard Const. & with Special tokens (with Recon.) &  18.83 & 11.84 &8.01 & 5.49  & 43.27\\
        Connection Const. & with Special tokens (with Recon.) &  19.17&11.90& 8.15& 5.68 &  45.85\\
        Katz Const.& with Special tokens (with Recon.) &  \textbf{19.30}&\textbf{12.10}& \textbf{8.30}& \textbf{5.83} &  \textbf{46.76}\\
        \bottomrule
    \end{tabular}
    \caption{\label{tab:ablation}Ablation study of each component in DialogGSR on OpenDialKG dataset.}
\end{table*}


%% file: Tables/llama_table.tex
\begin{table}
\centering

    \small
    \begin{tabular}{l|ccc}
        \toprule
        {Method} &  B-1 & B-2  \\
         \midrule 
         \midrule
         Base~(w/o KG) & 18.68 & 11.96 \\
         DialogGSR~(w/o Const.) & 19.60 & 13.32 \\
         DialogGSR~(ours) & \textbf{21.10} & \textbf{14.44} \\
        \bottomrule
    \end{tabular}
    \caption{Experimental results on OpenDialKG dataset with large language model \texttt{Llama-3-8b} under the fine-tuning with LoRA~{\cite{hu2022lora}}. `Const.' denotes graph-constrained decoding.
}
\label{tab:llm}

\end{table}

%% file: 4_Experiments/4_2_setup.tex
For fair comparisons with previous works, we use T5-small~\cite{roberts2019exploring} as the base PLM.
We select the best model on the validation set to evaluate the performance of all experiments.
More details are in Appendix~\ref{supp:exp_details}.

\paragraph{Evaluation metrics.}
We evaluate the dialog generation performance of different models with BLEU~\cite{papineni2002bleu}, ROUGE~\cite{lin2004rouge}, and unigram F1 score, by comparing the generated responses with the gold responses.
In addition, we use the KQA metric~\cite{kang2022knowledge}, which measures whether the factually correct and necessary knowledge is contained in the generated response given the dialog history. 
We also evaluate the performance of the retriever with path@k metrics, which are the recall@k of ground-truth paths following \cite{moon2019opendialkg,jung2020attnio}.

%% file: 4_Experiments/4_3_results.tex
We compare our DialogGSR with existing knowledge{\textendash}grounded dialog generation models on OpenDialKG dataset. 
Table~\ref{tab:opendialkg} shows that DialogGSR achieves the best performance in all metrics~(BLEU, ROUGE, KQA, and F1 score).
In particular, DialogGSR outperforms other baselines on KQA metrics by a large margin (4.61 on EM metric), which indicates that the proposed method generates more factually correct responses with relevant knowledge.
In addition, our method achieves a 1.53 performance gain on BLEU-1 metric compared to the best baseline method, which is an 8.61\% improvement.  
The performance gain of DialogGSR compared to SURGE, which retrieves the subgraph with a bi-encoder and uses graph neural networks for graph representations, indicates that our generative retrieval is effective in retrieving relevant knowledge and generating more accurate responses based on the retrieved knowledge. 

We also conduct experiments on KOMODIS~\cite{galetzka2020corpus} dataset.
Similar to the OpenDialKG result, Table~\ref{tab:komodis} demonstrates that our DialogGSR achieves the best performance compared to all the previous approaches.
To further validate the effectiveness of our generative subgraph retrieval, we compare the retrieval performance by path@k metrics.
Table~\ref{tab:retrieval} shows that DialogGSR achieves the best performance compared to the other baselines.  
This result indicates that our generative subgraph retrieval successfully retrieves context-relevant subgraphs from the knowledge graph by fully utilizing the power of pretrained language models. 

\subsection{Human Evaluation}
\label{sup_sec:human}
We conduct human evaluation to assess the generated responses of our dialog generation model.
The detailed process of human evaluation is in Appendix~\ref{sec:human_evaluation_detail}.
Table~\ref{tab:humaneval} shows the experimental results of the human evaluation, where DialogGSR outperforms SURGE in all the metrics (Consistency, Informativeness, Fluency). 
In particular, on the Consistency and Informativeness metrics, DialogGSR achieves statistically significant performance gains of 0.16 and 0.47 over SURGE (based on $t$-test with $p$-value $<0.05$), which indicates that our generative subgraph retrieval performs significantly better in retrieving informative knowledge compared to existing retrieval methods.
Our DialogGSR provides a relatively small performance gain of 0.11 on the Fluency metric.
Since the Fluency metric is more influenced by the language model's performance than the knowledge retrieval performance, it is reasonable to expect similar fluency scores when using the same base language model (T5-small) for fair comparisons.

%% file: 5_Analysis/5_0_analysis.tex
\input{Figure_tex/bottleneck}
\input{Tables/qualitative_analysis.tex}

We analyze DialogGSR to answer the following research questions: 
\textbf{[Q1]} Does each component of DialogGSR contribute to a performance improvement? 
\textbf{[Q2]} Are graph-constrained decoding and the entity informativeness score helpful for retrieving context-relevant subgraphs?
\textbf{[Q3]} Is GSR robust to the information bottleneck issue?
\textbf{[Q4]} Is DialogGSR effective with large language models (LLMs)?

\paragraph{Ablation studies.}
We provide the ablation studies to answer \textbf{[Q1], [Q2]} by empirically showing the contribution of each component of DialogGSR in Table~\ref{tab:ablation}.
\textbf{w/o Const.} is generative retrieval without graph-constrained decoding. 
\jyp{\textbf{Hard const.} is the retrieval with graph-constrained decoding but not considering entity informativeness score.}
\textbf{Connection} and \textbf{Katz} use entity informativeness scores based on Connection~($\mathcal{IS}_{\text{con}}$) and Katz metrics~($\mathcal{IS}_{\text{Katz}}$) referred in Section~\ref{sec:3.4}, respectively.
\textbf{with Special tokens (w/o Recon.)} uses special tokens to linearize the knowledge graph without graph reconstruction learning while \textbf{with Special tokens (w/ Recon.)} uses prompts learned with graph reconstruction.
Table~\ref{tab:ablation} shows that each component contributes to the performance improvement of the model.
In particular, graph-constrained decoding is crucial in our generative approach.

In addition, the models with graph constraints show improvements compared to the model without the constraints, which indicates that the graph constraint is important for the generative retrieval of knowledge subgraphs.
Also, using entity informative score~(Connection, Katz) performs better than graph constraints without it since the entity informativeness score reflects graph structural proximity in the decoding process.

\jyp{
\paragraph{Effectiveness of DialogGSR with LLMs.}
To assess the effectiveness of our DialogGSR with Large Language Models~(LLMs)~(\textbf{[Q4]}), we apply it to LLaMA-3~\cite{meta2024introducing}.
The experimental result is shown in Table~\ref{tab:llm}.
From the table, the performance gain of DialogGSR compared to the base model is 2.42 in BLEU-1 score.
In addition, the experimental result demonstrates that our proposed graph-constrained decoding is still important in LLMs. 
This indicates that DialogGSR is also effective in large language models.
}

\paragraph{Information bottleneck issue.}
\jyp{
Information bottleneck issue~\cite{humeau2020poly,lee2022generative} usually occurs when a long text sequence, such as a dialog history, is encoded into a single fixed length of vector.
To explore the robustness of DialogGSR to the information bottleneck issue~(\textbf{[Q3]}), we compare the retrieval performance of DialogGSR with the baselines such as DiffKG and SURGE with respect to the number of turns in dialog histories in Figure~\ref{fig:bottleneck}.
The result shows that DialogGSR is robust for long dialogs whereas the other methods often deteriorate as the number of turns increases.
}

\paragraph{Qualitative analysis.}
\jyp{
We perform qualitative analysis by comparing responses generated from SURGE and DialogGSR.
Table~\ref{tab:qual} shows a sampled \textbf{Gold response} and the responses generated by SURGE~(\textbf{Baseline response}) and DialogGSR~(\textbf{DialogGSR response}) given a multi-turn dialog.
From the table, DialogGSR retrieves more informative knowledge to generate responses compared to the baseline.
Given the last turn ``Do you by any chance remember who Mila Kunis is married too, I totally forgot”, DialogGSR successfully retrieves the knowledge information related to ‘Mila Kunis’ to help provide the appropriate response from the question while the baseline fails to retrieve information related to answer the question. In contrast, the baseline incorrectly retrieves knowledge information related to “Justin Timberlake”, who is mentioned in the past turn (4th turn), which results in a factually incorrect response. This demonstrates that generative retrieval is effective in retrieving informative knowledge and generating knowledge-grounded multi-turn dialogs.
More qualitative results are included in Appendix~\ref{supp:add_qual}.
}

%% file: Figure_tex/bottleneck.tex
\begin{figure}
\small
\centering
\includegraphics[width=0.8\linewidth]{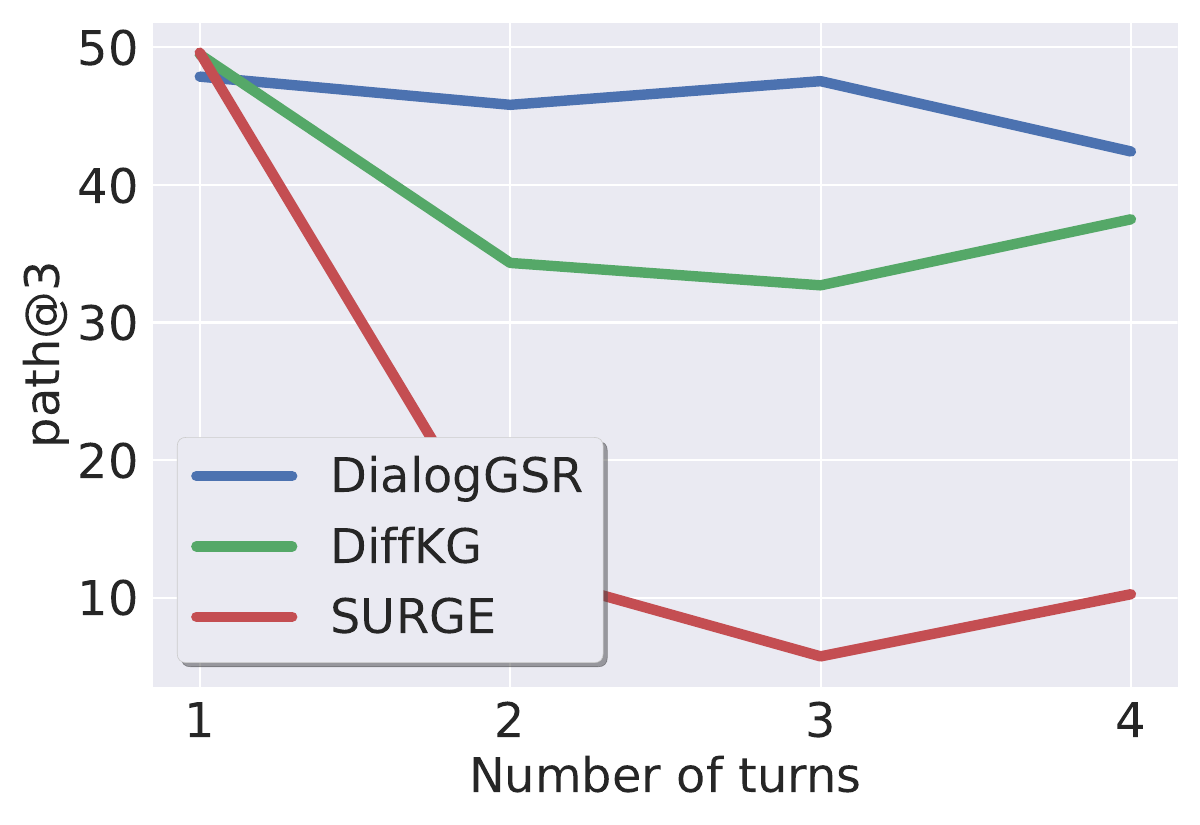}
\caption{Retrieval performance according to the number of turns.}
\label{fig:bottleneck}
\end{figure}

%% file: Tables/qualitative_analysis.tex
\begin{table*}[ht!]
\small
\centering
\resizebox{\textwidth}{!}  
{
    \begin{tabular}{llll}
        \toprule
          Dialog & Gold response & SURGE (Baseline) & DialogGSR (Ours) \\
         \midrule
         \midrule
         \begin{tabular}{@{}l@{}}
(a) Do you like Shaun White?\\
(b) I know he's an Olympic snowboarder he was funny in\\
\quad \ \ Friends With Benefits.\\
(a) Oh, I've never seen that movie, isn't Mila Kunis in it?\\ \quad \ \  I love her! \\
(b) She is. Justin Timberlake and Woody Harrelson were also \\ \quad \ \ in it. Shaun just played a small part. \\
(a) Do you by any chance remember who Mila Kunis is married \\
\quad \ \ too, I totally forgot.\\
\end{tabular}
& 
         \begin{tabular}{@{}l@{}}
She's married to \\ Ashton Kutcher.
\end{tabular} & 
         \begin{tabular}{@{}l@{}}
Mila Kunis is married \\ to Jennifer Lawrence.
\end{tabular} &
\begin{tabular}{@{}l@{}}
Mila Kunis is married \\ to Ashton Kutcher.
\end{tabular}\\         
    \midrule
        \multicolumn{4}{l}{\textit{\textbf{Knowledge triplets $\tau$ retrieved by Baseline}}} \\
        \multicolumn{4}{l}{$\langle$ `Justin Timberlake', `place musical career began', `Shelby Forest' $\rangle$} \\
        \multicolumn{4}{l}{$\langle$ `Justin Timberlake', `place musical career began', `Millington' $\rangle$} \\
        \multicolumn{4}{l}{$\langle$ `Justin Timberlake', `romantic relationship (with celebrities)', `Scarlett Johansson'$ \rangle$} \\
        \midrule
        \multicolumn{4}{l}{\textit{\textbf{Knowledge triplets $\tau$ retrieved by DialogGSR~(ours)}}}\\
        \multicolumn{4}{l}{$\langle$ `Ashton Kutcher', `romantic relationship (with celebrities)', `Mila Kunis' $\rangle$}\\
        \multicolumn{4}{l}{$\langle$ `Friends with Benefits', `starred\_actors', `Mila Kunis'  $\rangle$}\\
        \multicolumn{4}{l}{$\langle$ `Friends with Benefits', `starred\_actors', `Patricia Clarkson' $\rangle$}\\
\bottomrule
\end{tabular}
}
\caption{Comparison on responses generated by SURGE~(Baseline) and DialogGSR given a dialog.}
\label{tab:qual}
\end{table*}

%% file: 6_Conclusion/6_0_conclusion.tex
We have presented DialogGSR, a dialog generation model with generative subgraph retrieval. 
DialogGSR retrieves context-relevant subgraphs, by generating the subgraph token sequences considering both the dialog context and the graph information.
We have proposed novel knowledge graph linearization to convert knowledge triplets into token sequences with self-supervised graph-specific tokens to represent knowledge graphs without separate knowledge graph modules.
In addition, we have formulated a graph-constrained decoding for valid and relevant generative retrieval.
Our experiments demonstrate the effectiveness of our proposed method in knowledge\textendash{graph} grounded dialog generation.
Our codes are publicly available at \url{https://github.com/mlvlab/DialogGSR}.
\section*{Limitations}
The proposed DialogGSR generatively retrieves token sequences of the subgraph from a knowledge graph and then generates a response with the retrieved subgraph.
However, similar to works using graph retrieval on knowledge-grounded dialog generation, our generative subgraph retrieval can retrieve only the knowledge information contained in the knowledge graph.
Second, the benchmark datasets for knowledge graph{\textendash}grounded dialog generation are limited.
Therefore, new benchmark datasets on dialog generation with knowledge graphs warrants greater attention.

\section*{Ethics Statement}
Our DialogGSR does not have any direct negative social impacts, but it can potentially be used maliciously, similar to other dialog generation models. These models may produce factually incorrect or biased responses, particularly in sensitive areas such as politics, religion, and diplomacy. To address these risks, we advocate for the release of benchmark datasets without private information and emphasize the need for research into the methods that detect the source of texts. These measures are essential for the responsible development and use of dialog generation technologies.

%% file: appendix.tex
\clearpage

\section{Additional experiments}

\subsection{Additional Quantitative Analysis}
\label{sup_sec:aqa}
We also conduct experiments to verify the contribution of using $\texttt{[Rev]}$ to represent reverse relations and multiple consecutive tokens to represent each $\texttt{[Rev]}$ or $\texttt{[Int]}$ in Table~\ref{sup_tab:prompt}.
By adding \textbf{reverse tokens} to the knowledge, which allows mentioned entities that are tail entities in the provided triplets to be the starting points for the decoding, the performance is improved by 0.33 on BLEU-1 metric. 
Also, using \textbf{multiple consecutive tokens} to represent each $\texttt{[Rev]}$ or $\texttt{[Int]}$ (e.g., $\texttt{[Head]}$ $\Ent_1$ $\texttt{[Int{\textsubscript{\texttt{11}}}]}$ $\texttt{[Int{\textsubscript{\texttt{12}}}]}$ $\Rel_1$ $\texttt{[Int{\textsubscript{\texttt{21}}}]}$ $\texttt{[Int{\textsubscript{\texttt{22}}}]}$ $\Ent_{2}$ $\texttt{[Tail]}$) gives the performance gain on all the metrics since using the multiple tokens improve the capacity of representing the entities and the relations on top of language models.
By adding all the components, performance significantly improves by 0.56 on BLEU-1 metric compared to the linearized knowledge graph without any special tokens, which demonstrates the effectiveness of our proposed knowledge graph linearization approaches with special tokens. 
Interestingly, adding reverse tokens with using multiple consecutive tokens improves the overall performance compared to adding reverse tokens without using multiple consecutive tokens, which indicates that representing reverse relations is more effective when the capacity of the knowledge representation is increased. 

\input{Tables/supp_analysis_table}

\input{Tables/supp_qual_table}

\subsection{Additional Qualitative Analysis}
\label{supp:add_qual}
In Table~\ref{tab_sup:qual}, we provide additional qualitative examples for what we have shown in Table~\ref{tab:qual} of the main paper.
Our DialogGSR often generates high-quality responses similar to the main paper.
For example, in the first example, our DialogGSR generates a factually correct response "It was written by Frank Beddor" based on the retrieved triplet $\langle$`The Looking Glass Wars', `written\_by', `Frank Beddor'$\rangle$ while SURGE generates a factually incorrect response "Terry Pratchett" with the same triplet $\langle$`The Looking Glass Wars', `written\_by', `Frank Beddor'$\rangle$.
It demonstrates that our DialogGSR is more effective in generating responses even with the same knowledge information given. 
In the second example, DialogGSR successfully generates a factually correct response by retrieving context-relevant knowledge triplets whereas the factually incorrect response is generated by the baseline due to the retrieval of irrelevant knowledge.
These results demonstrate that our generative retrieval is effective in retrieving informative knowledge and generating knowledge-grounded dialogs.

\section{Experimental details}
\label{supp:exp_details}


\subsection{Implementation details}
In this section, we describe the implementation details not included in our main paper.
For all the experiments, we use PyTorch\footnote{Copyright (c) 2016-Facebook, Inc (Adam Paszke), Licensed under BSD-style license}~\cite{paszke2019pytorch} and Transformer module of Huggingface\footnote{Copyright 2018-The Hugging Face team, Licensed under the Apache license}~\cite{wolf2019huggingface} as our code base.
All experiments are conducted with 48GB NVIDIA RTX A6000 GPU.
We select the best model on the validation set to evaluate the performance of all experiments.
The epoch for training is set to 50 and the weight decay is 0.1. 
We use AdamW optimizer~\cite{Loshchilov2019adamw} to train our model and adopt learning rate decay.
\paragraph{Knowledge graph{\textendash}constrained decoding.}
Without the graph constraints, the language model is prone to generate invalid or irrelevant subgraphs due to the language model's bias~\cite{chen2022relation,de2021autoregressive}.
To inject the knowledge graph information into the language model in the decoding step, we present a knowledge graph{\textendash}constrained decoding method.
We use $\alpha=0.8$ and $k=2$ for calculating Katz~\cite{katz1953new} index-based entity informativeness score.
$p_{\text{graph}}$ is defined in Eq.~\eqref{eq:katz} of the main paper, and $b$ is 5.


\section{Details of Human Evaluation}
\label{sec:human_evaluation_detail}

We first randomly selected 30 dialogs from OpenDialKG test dataset~\cite{moon2019opendialkg} and generated responses using our model and SURGE~\cite{kang2022knowledge} for the comparison.
We recruited 22 participants who were not involved in our research and allowed the use of external sources, such as the Internet, to verify the factual correctness of generated responses. 
Following the process outlined in the other work~\cite{kang2022knowledge}, we utilized a 3-point Likert-like scale to evaluate three criteria: Consistency, Informativeness, and Fluency. 
\textbf{Consistency} measures the coherence and logical flow within the context of the conversation, \textbf{Informativeness} assesses the correctness and usefulness of the information in the generated responses, and \textbf{Fluency} focuses on the naturalness and linguistic quality of the dialog. 
With the human evaluation metrics and the automatic metrics in the main paper, we establish a comprehensive evaluation framework that enables accurate comparisons between models, enhancing the reliability of our assessment.

\input{Figure_tex/subgraph}

\section{Baselines}
\label{supp:baselines}
\subsection{Response Generation}
In our experiments, the following baseline models are used for comparing the response generation performance with our DialogGSR.

\begin{itemize}
    \item \textbf{T5{\textendash}small~(w/o KGs)}\footnote{Licensed under the Apache license}~\cite{roberts2019exploring}: T5-small is an encoder-decoder Transformer architecture for various natural language processing tasks.
    \item \textbf{Space Efficient (series)}\footnote{Copyright (c) 2021 Fabian Galetzka, Licensed under MIT license}~\cite{galetzka2021space}: Space Efficient (series) is the model proposed in \cite{galetzka2021space}. It utilizes all knowledge triplets related to the entities by matching the entities of KG and the entities mentioned in dialog history without any retrieval process.
    This model sequentially encodes knowledge triplets and feeds them into the encoder.
    \item \textbf{Space Efficient (parallel)}~\cite{galetzka2021space} : This model is also proposed by \cite{galetzka2021space}. Different from Space Efficient (series), this model constructs a segmentation block for each entity and encodes the relation in the segmentation block to reflect relational information.
    \item \textbf{Diff-KG}~\cite{tuan2022towards}: Diff-KG reasons differentiable knowledge paths to jointly generate a response with the dialog history.
    After the path reasoning, entities included in the path are concatenated with dialog history, and they are fed into a pretrained language model. 
    \item \textbf{SURGE (unsup.)}~\cite{kang2022knowledge}: SURGE is a graph neural network{\textendash}augmented Transformer-based dialog generation model that encodes knowledge triplets with graph neural networks.
    SURGE also retrieves context-relevant triplets via a subgraph retriever.
    This model trains the retriever without the guidance of gold knowledge and is implicitly trained with response generation loss.
    \item \textbf{SURGE (semi-sup.)}~\cite{kang2022knowledge}: SURGE (semi-sup.) uses gold knowledge to train the retriever. 
    \item \textbf{SURGE (contrastive)}~\cite{kang2022knowledge}: SURGE (contrastive) uses both the retrieval supervision from SURGE (Semi-sup.) and contrastive learning to encourage the encoder output and the decoder output to be closer.
\end{itemize}

\subsection{Knowledge Retrieval}
The models below are used as the baselines for validating the effectiveness of our DialogGSR on knowledge subgraph retrieval.

\begin{itemize}
    \item \textbf{Seq2Seq}~\cite{sutskever2014sequence}: Seq2Seq is used as a baseline in \cite{moon2019opendialkg,tuan2022towards}. Given all of the dialog contexts, Seq2Seq generates entity paths.
    \item \textbf{Tri-LSTM}~\cite{young2017augmenting}: Tri-LSTM is another baseline in \cite{moon2019opendialkg,tuan2022towards}. It encodes dialog contexts and related 1-hop knowledge from a KG to retrieve knowledge paths. 
    \item \textbf{Ext-ED (Extended Encoder-Decoder)}~\cite{parthasarathi2018extending}: Extended Encoder-Decoder is also one of the baselines in \cite{moon2019opendialkg,tuan2022towards}. 
    It generates a response conditioned on an external knowledge vector input, which is encoded by GloVe embedding.
    \item \textbf{DialKG Walker}~\cite{moon2019opendialkg}: DialKG Walker is an attention-based knowledge path retrieval model designed to traverse a knowledge graph with dialog context and knowledge paths.
    \item \textbf{AttnFlow}~\cite{jung2020attnio}: AttnFlow is an attention-based knowledge path retrieval model based on GAT~\cite{velivckovic2018graph} and the encoded dialog context.
    It only uses incoming attention flow to update knowledge representation.
    \item \textbf{AttnIO}~\cite{jung2020attnio}: AttnIO is an extension of AttnFlow, where both incoming and outcoming attention flows are used to represent knowledge paths with dialog contexts and entity features.
\end{itemize}

%% file: Tables/supp_analysis_table.tex
\begin{table}[t]
    \small
    \centering
    
    \begin{tabular}{cc|ccccc}
        \toprule
	 Reverse & Multiple &   {B-1} & {B-2} & path@3\\
         \midrule 
         \midrule
            &            & 18.74&11.99&   {41.28}\\
        \checkmark &            & 19.07&12.03&  {44.54}\\
                   & \checkmark & 19.22&12.01&   {45.06}\\
        \checkmark & \checkmark & \textbf{19.30}&\textbf{12.10} &  \textbf{46.76}\\
        \bottomrule
    \end{tabular}
    \caption{Ablation studies on special tokens with OpenDialKG dataset. `Reverse' denotes reverse tokens and `Multiple' denotes multiple tokens.}
    \label{sup_tab:prompt}
\end{table}

%% file: Tables/supp_qual_table.tex
\begin{table*}[t]
    \centering    
    \resizebox{\textwidth}{!}    
    {
    \begin{tabular}{llll}
        \toprule
          Dialog & Gold response & SURGE (Baseline) & DialogGSR (Ours) \\
         \midrule
         \midrule
         \begin{tabular}{@{}l@{}}
(a) Could you recommend and books by the author of\\ \quad \ \  Colour of Magic? \\
(b) The Colour of Magic has genre fantasy. So do you\\ \quad \ \  want to read fantasy books? \\
(a) Like Through  the Looking Glass?  Sure I like \\ \ \ \quad  Fantasy okay. \\
(b) yes like The Looking Glass Wars it's really a good \\ \quad \ \  book. I suggest reading it. \\
(a) Do you know who wrote it by any chance? \\
\end{tabular}
& 
         \begin{tabular}{@{}l@{}}
Yes Frank Beddor wrote it,\\ who also wrote Seeing Redd.
\end{tabular} & 
         \begin{tabular}{@{}l@{}}
Terry Pratchett
\end{tabular} &
\begin{tabular}{@{}l@{}}
It was written by \\Frank Beddor.
\end{tabular}\\         
    \midrule
        \multicolumn{4}{l}{\textit{\textbf{Knowledge triplets $\tau$ retrieved from Baseline}}} \\
        \multicolumn{4}{l}{$\langle$ `The Colour of Magic', `written\_by', `Terry Pratchett' $\rangle$} \\
        \multicolumn{4}{l}{$\langle$ `The Looking Glass Wars', `written\_by', `Frank Beddor' $\rangle$} \\
        \midrule
        \multicolumn{4}{l}{\textit{\textbf{Knowledge triplets $\tau$ retrieved from DialogGSR}}}\\
        \multicolumn{4}{l}{$\langle$ `The Looking Glass Wars', `written\_by', `Frank Beddor' $\rangle$} \\
        \multicolumn{4}{l}{$\langle$ `Frank Beddor', `is-a', `Film Producer' $\rangle$} \\
        \midrule
        \midrule
         \begin{tabular}{@{}l@{}}
(a) I like the book Where'd You Go, Bernadette. \\
\quad \ \ Do you have any other suggestions for me? \\
(b) Definitely!  That's a great book by Maria Semple.\\ \quad \ \   Do you like her? \\
(a) I do! Has she written anything else? \\
\end{tabular}
& 
         \begin{tabular}{@{}l@{}}
She is a screenwriter,\\ television producer, and\\ she produced Mad About You.

\end{tabular} & 
         \begin{tabular}{@{}l@{}}
She's written a lot \\of books, including \\Where'd You Go, Bernadette.\\ Have you read that one?
\end{tabular} &
\begin{tabular}{@{}l@{}}
She has. She also \\wrote the TV program,\\ Mad About You.\\ Have you heard of that one?
\end{tabular}\\         
    \midrule
        \multicolumn{4}{l}{\textit{\textbf{Knowledge triplets $\tau$ retrieved from Baseline}}} \\
        \multicolumn{4}{l}{$\langle$ `Where'd You Go, Bernadette', `written\_by',	`Maria Semple' $\rangle$} \\
        \multicolumn{4}{l}{$\langle$ `Where'd You Go, Bernadette',	`release\_year',	`2012' $\rangle$} \\
        \multicolumn{4}{l}{$\langle$ `2012',	`release\_year (reverse)',	`Where'd You Go, Bernadette' $\rangle$} \\
        \midrule
        \multicolumn{4}{l}{\textit{\textbf{Knowledge triplets $\tau$ retrieved from DialogGSR}}}\\
        \multicolumn{4}{l}{$\langle$ `Where'd You Go, Bernadette', `written\_by' ,`Maria Semple' $\rangle$} \\
        \multicolumn{4}{l}{$\langle$ `Maria Semple', `tv program produced', `Mad About You' $\rangle$} \\        
        \multicolumn{4}{l}{$\langle$ `Mad About You', `has\_genre', `sitcom' $\rangle$} \\

        \midrule
        \midrule
         \begin{tabular}{@{}l@{}}
(a) Do you know any movies directed by \\ \quad \ \  Bennett Miller? \\
(b) He has some great ones. Have you seen \\ \quad \ \ Moneyball or Capote? \\
(a) I haven't seen Moneyball,\\ \quad \ \  who stars in it? \\
\end{tabular}
& 
         \begin{tabular}{@{}l@{}}
Steve Zaillian wrote Moneyball.\\ It starred Brad Pitt along with\\ Tammy Blanchard. It's a \\ really good movie!

\end{tabular} & 
         \begin{tabular}{@{}l@{}}
Capote stars Seymour\\ Hoffman and Ben\\ Stiller. It's a \\ romantic comedy.

\end{tabular} &
\begin{tabular}{@{}l@{}}
Tammy BLanchard and \\Brad  Pitt are in it. \\Do you like action movies?
\end{tabular}\\         
    \midrule
        \multicolumn{4}{l}{\textit{\textbf{Knowledge triplets $\tau$ retrieved from Baseline}}} \\
        \multicolumn{4}{l}{$\langle$ `Moneyball', `starred\_actors', `Philip Seymour Hoffman' $\rangle$} \\
        \multicolumn{4}{l}{$\langle$ `Capote', `starred\_actors',`Philip Seymour Hoffman' $\rangle$} \\
        \multicolumn{4}{l}{$\langle$ `Philip Seymour Hoffman', `starred\_actors (reverse)', `Moneyball' $\rangle$} \\
        \midrule
        \multicolumn{4}{l}{\textit{\textbf{Knowledge triplets $\tau$ retrieved from DialogGSR}}}\\
        \multicolumn{4}{l}{$\langle$ `Moneyball', `starred\_actors', `Tammy Blanchard' $\rangle$} \\
        \multicolumn{4}{l}{$\langle$ `Moneyball', `starred\_actors', `Brad Pitt' $\rangle$} \\        
        \multicolumn{4}{l}{$\langle$ `Moneyball', `starred\_actors', `Robin Wright' $\rangle$} \\

        \midrule
        \midrule
         \begin{tabular}{@{}l@{}}
(a) Can you tell me some \\ \quad \ \  Beatrix Potter books? \\
(b) Would you prefer her biography,\\ \quad \ \  or children's literature? \\
(a) Children's literature please. \\
(b) Would you like The Tale of Peter \\ \quad \ \  Rabbit? Or The Tale of Mrs Tiggy-Winkle? \\ \quad \ \  I hear they're both good.\\
(a) Ill try Peter Rabbit,\\ \quad \ \  out of curiosity when \\ \quad \ \ were these books written? \\
\end{tabular}
& 
         \begin{tabular}{@{}l@{}}
I believe they were written \\ in the year 1905, to be exact.

\end{tabular} & 
         \begin{tabular}{@{}l@{}}
The Tale of Mrs. Tiggy-Winkle \\ was written in 1851.

\end{tabular} &
\begin{tabular}{@{}l@{}}
I'm not sure when they \\ were written, but they were\\ released in the year 1905.
\end{tabular}\\         
    \midrule
        \multicolumn{4}{l}{\textit{\textbf{Knowledge triplets $\tau$ retrieved from Baseline}}} \\
        \multicolumn{4}{l}{$\langle$ `The Complete Tales of Beatrix Potter', `written\_by', `Beatrix Potter
' $\rangle$} \\
        \multicolumn{4}{l}{$\langle$ `The Complete Adventures of Peter Rabbit', `written\_by', `Beatrix Potter
' $\rangle$} \\
        \multicolumn{4}{l}{$\langle$ `The Tale of Mrs. Tiggy-Winkle', `written\_by', `Beatrix Potter
' $\rangle$} \\
        \midrule
        \multicolumn{4}{l}{\textit{\textbf{Knowledge triplets $\tau$ retrieved from DialogGSR}}}\\
        \multicolumn{4}{l}{$\langle$ `The Tale of Mrs. Tiggy-Winkle', `written\_by', `Beatrix Potter' $\rangle$} \\
        \multicolumn{4}{l}{$\langle$ `The Tale of Mrs. Tiggy-Winkle', `release\_year', `1905' $\rangle$} \\        
        \multicolumn{4}{l}{$\langle$ `The Return Of Sherlock Holmes', `release\_year', `1905' $\rangle$} \\
      
    \bottomrule
    \end{tabular}
    }
    \caption{\label{tab_sup:qual}Comparison on responses generated by SURGE~(Baseline) and DialogGSR given a dialog.}
\end{table*}

%% file: Figure_tex/subgraph.tex
\begin{figure}

\centering
\includegraphics[width=1.0\linewidth]{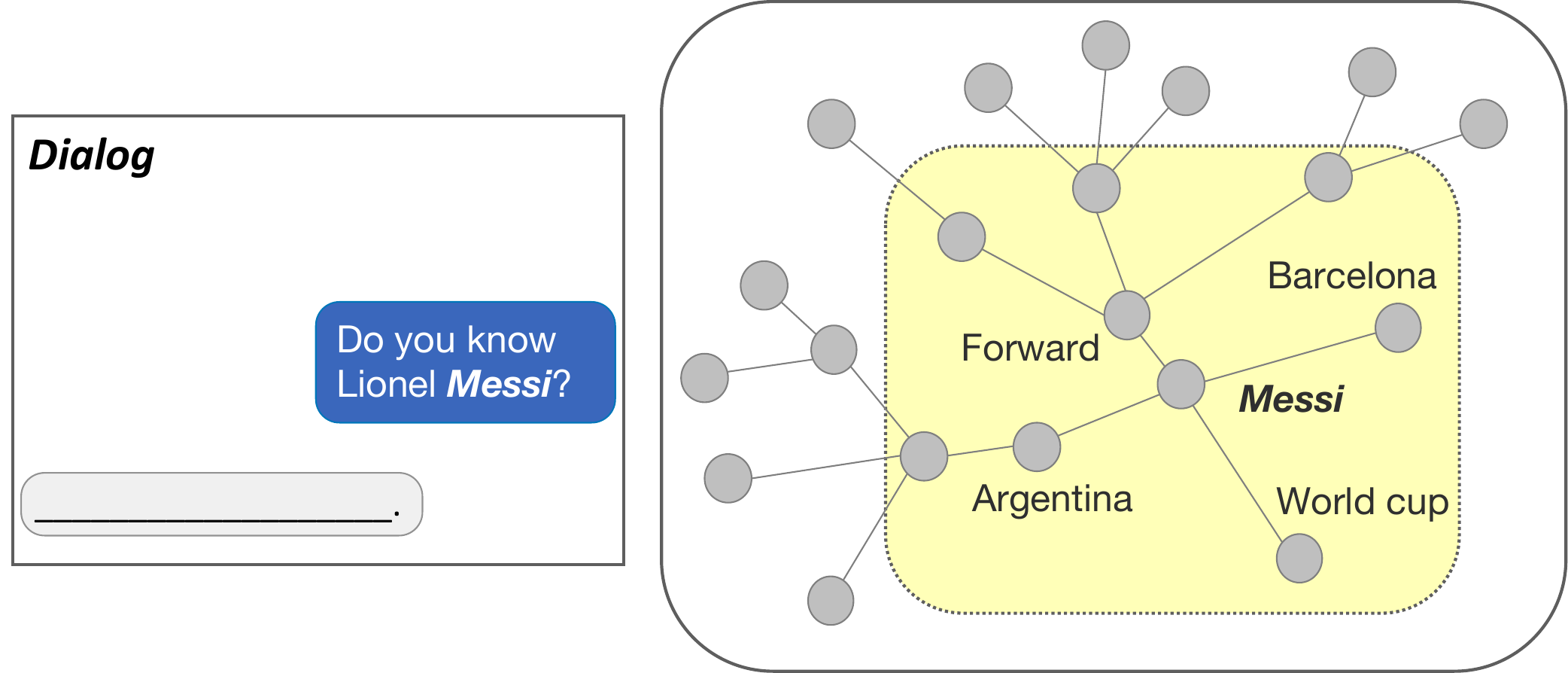}
\caption{An example of extracting a 2-hop candidate subgraph from the knowledge graph. Yellow region indicates the 2-hop candidate subgraph centered on the mentioned entity ``Messi".}
\label{fig:subgraph}
\end{figure}

%% file: main_cam.bbl
\begin{thebibliography}{72}
\expandafter\ifx\csname natexlab\endcsname\relax\def\natexlab#1{#1}\fi

\bibitem[{Achiam et~al.(2023)Achiam, Adler, Agarwal, Ahmad, Akkaya, Aleman, Almeida, Altenschmidt, Altman, Anadkat et~al.}]{achiam2023gpt}
Josh Achiam, Steven Adler, Sandhini Agarwal, Lama Ahmad, Ilge Akkaya, Florencia~Leoni Aleman, Diogo Almeida, Janko Altenschmidt, Sam Altman, Shyamal Anadkat, et~al. 2023.
\newblock Gpt-4 technical report.
\newblock \emph{arXiv:2303.08774}.

\bibitem[{Balakrishnan et~al.(2019)Balakrishnan, Rao, Upasani, White, and Subba}]{balakrishnan2019constrained}
Anusha Balakrishnan, Jinfeng Rao, Kartikeya Upasani, Michael White, and Rajen Subba. 2019.
\newblock Constrained decoding for neural {NLG} from compositional representations in task-oriented dialogue.
\newblock In \emph{ACL}, pages 832--844.

\bibitem[{Bast et~al.(2014)Bast, B{\"a}urle, Buchhold, and Hau{\ss}mann}]{bast2014easy}
Hannah Bast, Florian B{\"a}urle, Bj{\"o}rn Buchhold, and Elmar Hau{\ss}mann. 2014.
\newblock Easy access to the freebase dataset.
\newblock In \emph{WWW}, pages 95--98.

\bibitem[{Bevilacqua et~al.(2022)Bevilacqua, Ottaviano, Lewis, Yih, Riedel, and Petroni}]{bevilacqua2022autoregressive}
Michele Bevilacqua, Giuseppe Ottaviano, Patrick Lewis, Scott Yih, Sebastian Riedel, and Fabio Petroni. 2022.
\newblock Autoregressive search engines: Generating substrings as document identifiers.
\newblock In \emph{NeurIPS}, pages 31668--31683.

\bibitem[{Bordes et~al.(2013)Bordes, Usunier, Garcia-Duran, Weston, and Yakhnenko}]{NIPS2013_1cecc7a7}
Antoine Bordes, Nicolas Usunier, Alberto Garcia-Duran, Jason Weston, and Oksana Yakhnenko. 2013.
\newblock Translating embeddings for modeling multi-relational data.
\newblock In \emph{NeurIPS}, pages 2787--2795.

\bibitem[{Brin(1998)}]{brin1998pagerank}
Sergey Brin. 1998.
\newblock The pagerank citation ranking: bringing order to the web.
\newblock \emph{Proceedings of ASIS, 1998}, 98:161--172.

\bibitem[{Cao et~al.(2021)Cao, Izacard, Riedel, and Petroni}]{de2021autoregressive}
Nicola~De Cao, Gautier Izacard, Sebastian Riedel, and Fabio Petroni. 2021.
\newblock Autoregressive entity retrieval.
\newblock In \emph{ICLR}.

\bibitem[{Chen et~al.(2017)Chen, Fisch, Weston, and Bordes}]{ChenFWB17}
Danqi Chen, Adam Fisch, Jason Weston, and Antoine Bordes. 2017.
\newblock Reading wikipedia to answer open-domain questions.
\newblock In \emph{ACL}, pages 1870--1879.

\bibitem[{Chen et~al.(2022{\natexlab{a}})Chen, Zhang, Guo, Fan, and Cheng}]{chen2022gere}
Jiangui Chen, Ruqing Zhang, Jiafeng Guo, Yixing Fan, and Xueqi Cheng. 2022{\natexlab{a}}.
\newblock Gere: Generative evidence retrieval for fact verification.
\newblock In \emph{SIGIR}, pages 2184--2189.

\bibitem[{Chen et~al.(2022{\natexlab{b}})Chen, Yang, and Wan}]{chen2022relation}
Xiang Chen, Zhixian Yang, and Xiaojun Wan. 2022{\natexlab{b}}.
\newblock Relation-constrained decoding for text generation.
\newblock In \emph{NeurIPS}, pages 26804--26819.

\bibitem[{Devlin et~al.(2019)Devlin, Chang, Lee, and Toutanova}]{devlin2018bert}
Jacob Devlin, Ming-Wei Chang, Kenton Lee, and Kristina Toutanova. 2019.
\newblock {BERT}: Pre-training of deep bidirectional transformers for language understanding.
\newblock In \emph{NAACL-HLT}, pages 4171--4186.

\bibitem[{Dinan et~al.(2019)Dinan, Roller, Shuster, Fan, Auli, and Weston}]{dinan2019wizard}
Emily Dinan, Stephen Roller, Kurt Shuster, Angela Fan, Michael Auli, and Jason Weston. 2019.
\newblock Wizard of wikipedia: Knowledge-powered conversational agents.
\newblock In \emph{ICLR}.

\bibitem[{Du{\v{s}}ek et~al.(2018)Du{\v{s}}ek, Novikova, and Rieser}]{duvsek2018findings}
Ond{\v{r}}ej Du{\v{s}}ek, Jekaterina Novikova, and Verena Rieser. 2018.
\newblock Findings of the {E2E NLG} challenge.
\newblock In \emph{INLG}, pages 322--328.

\bibitem[{Du{\v{s}}ek et~al.(2020)Du{\v{s}}ek, Novikova, and Rieser}]{duvsek2020evaluating}
Ond{\v{r}}ej Du{\v{s}}ek, Jekaterina Novikova, and Verena Rieser. 2020.
\newblock Evaluating the state-of-the-art of end-to-end natural language generation: The {E2E NLG} challenge.
\newblock \emph{Computer Speech \& Language}, 59:123--156.

\bibitem[{Feng et~al.(2020)Feng, Chen, Lin, Wang, Yan, and Ren}]{feng2020scalable}
Yanlin Feng, Xinyue Chen, Bill~Yuchen Lin, Peifeng Wang, Jun Yan, and Xiang Ren. 2020.
\newblock Scalable multi-hop relational reasoning for knowledge-aware question answering.
\newblock In \emph{EMNLP}, pages 1295--1309.

\bibitem[{Fernandes et~al.(2019)Fernandes, Allamanis, and Brockschmidt}]{fernandes2018structured}
Patrick Fernandes, Miltiadis Allamanis, and Marc Brockschmidt. 2019.
\newblock Structured neural summarization.
\newblock In \emph{ICLR}.

\bibitem[{Galetzka et~al.(2020)Galetzka, Eneh, and Schlangen}]{galetzka2020corpus}
Fabian Galetzka, Chukwuemeka~U Eneh, and David Schlangen. 2020.
\newblock A corpus of controlled opinionated and knowledgeable movie discussions for training neural conversation models.
\newblock In \emph{LREC}, pages 565--573.

\bibitem[{Galetzka et~al.(2021)Galetzka, Rose, Schlangen, and Lehmann}]{galetzka2021space}
Fabian Galetzka, Jewgeni Rose, David Schlangen, and Jens Lehmann. 2021.
\newblock Space efficient context encoding for non-task-oriented dialogue generation with graph attention transformer.
\newblock In \emph{ACL-IJCNLP}, pages 7028--7041.

\bibitem[{Gasteiger et~al.(2019)Gasteiger, Bojchevski, and G{\"u}nnemann}]{gasteiger2018predict}
Johannes Gasteiger, Aleksandar Bojchevski, and Stephan G{\"u}nnemann. 2019.
\newblock Predict then propagate: Graph neural networks meet personalized pagerank.
\newblock In \emph{ICLR}.

\bibitem[{Ghazvininejad et~al.(2018)Ghazvininejad, Brockett, Chang, Dolan, Gao, Yih, and Galley}]{ghazvininejad2018knowledge}
Marjan Ghazvininejad, Chris Brockett, Ming-Wei Chang, Bill Dolan, Jianfeng Gao, Wen-tau Yih, and Michel Galley. 2018.
\newblock A knowledge-grounded neural conversation model.
\newblock In \emph{AAAI}, pages 5110--5117.

\bibitem[{Hu et~al.(2022)Hu, Shen, Wallis, Allen-Zhu, Li, Wang, Wang, and Chen}]{hu2022lora}
Edward~J Hu, Yelong Shen, Phillip Wallis, Zeyuan Allen-Zhu, Yuanzhi Li, Shean Wang, Lu~Wang, and Weizhu Chen. 2022.
\newblock Lo{RA}: Low-rank adaptation of large language models.
\newblock In \emph{ICLR}.

\bibitem[{Huang et~al.(2020)Huang, Wu, and Wang}]{huang2020knowledge}
Luyang Huang, Lingfei Wu, and Lu~Wang. 2020.
\newblock Knowledge graph-augmented abstractive summarization with semantic-driven cloze reward.
\newblock In \emph{ACL}, pages 5094--5107.

\bibitem[{Humeau et~al.(2020)Humeau, Shuster, Lachaux, and Weston}]{humeau2020poly}
Samuel Humeau, Kurt Shuster, Marie-Anne Lachaux, and Jason Weston. 2020.
\newblock Poly-encoders: Transformer architectures and pre-training strategies for fast and accurate multi-sentence scoring.
\newblock In \emph{ICLR}.

\bibitem[{Izacard and Grave(2021)}]{IzacardG21}
Gautier Izacard and Edouard Grave. 2021.
\newblock Leveraging passage retrieval with generative models for open domain question answering.
\newblock In \emph{EACL}, pages 874--880.

\bibitem[{Izacard et~al.(2020)Izacard, Petroni, Hosseini, De~Cao, Riedel, and Grave}]{izacard2020memory}
Gautier Izacard, Fabio Petroni, Lucas Hosseini, Nicola De~Cao, Sebastian Riedel, and Edouard Grave. 2020.
\newblock A memory efficient baseline for open domain question answering.
\newblock \emph{arXiv:2012.15156}.

\bibitem[{Ji et~al.(2023)Ji, Liu, Lee, Yu, Wilie, Zeng, and Fung}]{ji2022rho}
Ziwei Ji, Zihan Liu, Nayeon Lee, Tiezheng Yu, Bryan Wilie, Min Zeng, and Pascale Fung. 2023.
\newblock {RHO:} reducing hallucination in open-domain dialogues with knowledge grounding.
\newblock In \emph{ACL-findings}, pages 4504--4522.

\bibitem[{Jung et~al.(2020)Jung, Son, and Lyu}]{jung2020attnio}
Jaehun Jung, Bokyung Son, and Sungwon Lyu. 2020.
\newblock Attnio: Knowledge graph exploration with in-and-out attention flow for knowledge-grounded dialogue.
\newblock In \emph{EMNLP}, pages 3484--3497.

\bibitem[{Kang et~al.(2023)Kang, Kwak, Baek, and Hwang}]{kang2022knowledge}
Minki Kang, Jin~Myung Kwak, Jinheon Baek, and Sung~Ju Hwang. 2023.
\newblock Knowledge graph-augmented language models for knowledge-grounded dialogue generation.
\newblock \emph{arXiv:2305.18846}.

\bibitem[{Katz(1953)}]{katz1953new}
Leo Katz. 1953.
\newblock A new status index derived from sociometric analysis.
\newblock \emph{Psychometrika}, 18(1):39--43.

\bibitem[{Lee et~al.(2023)Lee, Kim, Chang, Oh, Yang, Karpukhin, Lu, and Seo}]{lee2023contextualized}
Hyunji Lee, Jaeyoung Kim, Hoyeon Chang, Hanseok Oh, Sohee Yang, Vlad Karpukhin, Yi~Lu, and Minjoon Seo. 2023.
\newblock Nonparametric decoding for generative retrieval.
\newblock In \emph{ACL-findings}, pages 12642--12661.

\bibitem[{Lee et~al.(2022)Lee, Yang, Oh, and Seo}]{lee2022generative}
Hyunji Lee, Sohee Yang, Hanseok Oh, and Minjoon Seo. 2022.
\newblock Generative multi-hop retrieval.
\newblock In \emph{EMNLP}, pages 1417--1436.

\bibitem[{Lewis et~al.(2020)Lewis, Perez, Piktus, Petroni, Karpukhin, Goyal, K{\"u}ttler, Lewis, Yih, Rockt{\"a}schel et~al.}]{lewis2020retrieval}
Patrick Lewis, Ethan Perez, Aleksandra Piktus, Fabio Petroni, Vladimir Karpukhin, Naman Goyal, Heinrich K{\"u}ttler, Mike Lewis, Wen-tau Yih, Tim Rockt{\"a}schel, et~al. 2020.
\newblock Retrieval-augmented generation for knowledge-intensive {NLP} tasks.
\newblock In \emph{NeurIPS}, pages 9459--9474.

\bibitem[{Lian et~al.(2019)Lian, Xie, Wang, Peng, and Wu}]{lian2019learning}
Rongzhong Lian, Min Xie, Fan Wang, Jinhua Peng, and Hua Wu. 2019.
\newblock Learning to select knowledge for response generation in dialog systems.
\newblock In \emph{IJCAI}, pages 5081--5087.

\bibitem[{Lin(2004)}]{lin2004rouge}
Chin-Yew Lin. 2004.
\newblock {ROUGE}: A package for automatic evaluation of summaries.
\newblock In \emph{Text Summarization Branches Out}, pages 74--81.

\bibitem[{Loshchilov and Hutter(2019)}]{Loshchilov2019adamw}
Ilya Loshchilov and Frank Hutter. 2019.
\newblock Decoupled weight decay regularization.
\newblock In \emph{ICLR}.

\bibitem[{Luan et~al.(2021)Luan, Eisenstein, Toutanova, and Collins}]{luan2021sparse}
Yi~Luan, Jacob Eisenstein, Kristina Toutanova, and Michael Collins. 2021.
\newblock Sparse, dense, and attentional representations for text retrieval.
\newblock \emph{TACL}, 9:329--345.

\bibitem[{Luo et~al.(2024)Luo, Li, Haffari, and Pan}]{luo2024reasoning}
Linhao Luo, Yuan-Fang Li, Gholamreza Haffari, and Shirui Pan. 2024.
\newblock Reasoning on graphs: Faithful and interpretable large language model reasoning.
\newblock In \emph{ICLR}.

\bibitem[{Meta(2024)}]{meta2024introducing}
AI~Meta. 2024.
\newblock Introducing meta llama 3: The most capable openly available llm to date.
\newblock \emph{Meta AI.}

\bibitem[{Moon et~al.(2019)Moon, Shah, Kumar, and Subba}]{moon2019opendialkg}
Seungwhan Moon, Pararth Shah, Anuj Kumar, and Rajen Subba. 2019.
\newblock {O}pen{D}ial{KG}: Explainable conversational reasoning with attention-based walks over knowledge graphs.
\newblock In \emph{ACL}, pages 845--854.

\bibitem[{Papineni et~al.(2002)Papineni, Roukos, Ward, and Zhu}]{papineni2002bleu}
Kishore Papineni, Salim Roukos, Todd Ward, and Wei-Jing Zhu. 2002.
\newblock {BLEU}: a method for automatic evaluation of machine translation.
\newblock In \emph{ACL}, pages 311--318.

\bibitem[{Park et~al.(2023)Park, Choi, Ko, Park, Kim, Jeong, Kim, and Kim}]{park2023relation}
Jinyoung Park, Hyeong~Kyu Choi, Juyeon Ko, Hyeonjin Park, Ji-Hoon Kim, Jisu Jeong, Kyungmin Kim, and Hyunwoo Kim. 2023.
\newblock Relation-aware language-graph transformer for question answering.
\newblock In \emph{AAAI}, pages 13457--13464.

\bibitem[{Parthasarathi and Pineau(2018)}]{parthasarathi2018extending}
Prasanna Parthasarathi and Joelle Pineau. 2018.
\newblock Extending neural generative conversational model using external knowledge sources.
\newblock In \emph{EMNLP}, pages 690--695.

\bibitem[{Paszke et~al.(2019)Paszke, Gross, Massa, Lerer, Bradbury, Chanan, Killeen, Lin, Gimelshein, Antiga et~al.}]{paszke2019pytorch}
Adam Paszke, Sam Gross, Francisco Massa, Adam Lerer, James Bradbury, Gregory Chanan, Trevor Killeen, Zeming Lin, Natalia Gimelshein, Luca Antiga, et~al. 2019.
\newblock Pytorch: An imperative style, high-performance deep learning library.
\newblock In \emph{NeurIPS}, pages 8024--8035.

\bibitem[{Qi et~al.(2023)Qi, Li, Hui, Wang, Li, Wu, and Laili}]{qi2023investigation}
Chengwen Qi, Bowen Li, Binyuan Hui, Bailin Wang, Jinyang Li, Jinwang Wu, and Yuanjun Laili. 2023.
\newblock An investigation of llms' inefficacy in understanding converse relations.
\newblock In \emph{EMNLP}, pages 6932--6953.

\bibitem[{Radford et~al.(2019)Radford, Wu, Child, Luan, Amodei, Sutskever et~al.}]{radford2019language}
Alec Radford, Jeffrey Wu, Rewon Child, David Luan, Dario Amodei, Ilya Sutskever, et~al. 2019.
\newblock Language models are unsupervised multitask learners.
\newblock \emph{OpenAI blog}, 1(8):9.

\bibitem[{Roberts et~al.(2020)Roberts, Raffel, Lee, Matena, Shazeer, Liu, Narang, Li, and Zhou}]{roberts2019exploring}
Adam Roberts, Colin Raffel, Katherine Lee, Michael Matena, Noam Shazeer, Peter~J Liu, Sharan Narang, Wei Li, and Yanqi Zhou. 2020.
\newblock Exploring the limits of transfer learning with a unified text-to-text transformer.
\newblock \emph{JMLR}, 21:140:1--140:67.

\bibitem[{Shuster et~al.(2021)Shuster, Poff, Chen, Kiela, and Weston}]{shuster2021retrieval}
Kurt Shuster, Spencer Poff, Moya Chen, Douwe Kiela, and Jason Weston. 2021.
\newblock Retrieval augmentation reduces hallucination in conversation.
\newblock In \emph{EMNLP-findings}, pages 3784--3803.

\bibitem[{Sun et~al.(2023)Sun, Ren, and Ren}]{sun2023generative}
Weiwei Sun, Pengjie Ren, and Zhaochun Ren. 2023.
\newblock Generative knowledge selection for knowledge-grounded dialogues.
\newblock In \emph{EACL-findings}, pages 2032--2043.

\bibitem[{Sutskever et~al.(2014)Sutskever, Vinyals, and Le}]{sutskever2014sequence}
Ilya Sutskever, Oriol Vinyals, and Quoc~V Le. 2014.
\newblock Sequence to sequence learning with neural networks.
\newblock In \emph{NeurIPS}, pages 3104--3112.

\bibitem[{Thoppilan et~al.(2022)Thoppilan, De~Freitas, Hall, Shazeer, Kulshreshtha, Cheng, Jin, Bos, Baker, Du et~al.}]{thoppilan2022lamda}
Romal Thoppilan, Daniel De~Freitas, Jamie Hall, Noam Shazeer, Apoorv Kulshreshtha, Heng-Tze Cheng, Alicia Jin, Taylor Bos, Leslie Baker, Yu~Du, et~al. 2022.
\newblock {LaMDA}: Language models for dialog applications.
\newblock \emph{arXiv:2201.08239}.

\bibitem[{Thorne(2022)}]{thorne2022data}
James Thorne. 2022.
\newblock Data-efficient auto-regressive document retrieval for fact verification.
\newblock In \emph{SustaiNLP}, pages 44--51.

\bibitem[{Thorne et~al.(2018)Thorne, Vlachos, Christodoulopoulos, and Mittal}]{ThorneVCM18}
James Thorne, Andreas Vlachos, Christos Christodoulopoulos, and Arpit Mittal. 2018.
\newblock {FEVER:} a large-scale dataset for fact extraction and verification.
\newblock In \emph{NAACL}, pages 809--819.

\bibitem[{Touvron et~al.(2023)Touvron, Martin, Stone, Albert, Almahairi, Babaei, Bashlykov, Batra, Bhargava, Bhosale et~al.}]{touvron2023llama}
Hugo Touvron, Louis Martin, Kevin Stone, Peter Albert, Amjad Almahairi, Yasmine Babaei, Nikolay Bashlykov, Soumya Batra, Prajjwal Bhargava, Shruti Bhosale, et~al. 2023.
\newblock Llama 2: Open foundation and fine-tuned chat models.
\newblock \emph{arXiv:2307.09288}.

\bibitem[{Tuan et~al.(2022)Tuan, Beygi, Fazel-Zarandi, Gao, Cervone, and Wang}]{tuan2022towards}
Yi-Lin Tuan, Sajjad Beygi, Maryam Fazel-Zarandi, Qiaozi Gao, Alessandra Cervone, and William~Yang Wang. 2022.
\newblock Towards large-scale interpretable knowledge graph reasoning for dialogue systems.
\newblock In \emph{ACL-findings}, pages 383--395.

\bibitem[{Tuan et~al.(2019)Tuan, Chen, and Lee}]{tuan2019dykgchat}
Yi-Lin Tuan, Yun-Nung Chen, and Hung-yi Lee. 2019.
\newblock {D}y{K}g{C}hat: Benchmarking dialogue generation grounding on dynamic knowledge graphs.
\newblock In \emph{EMNLP}, pages 1855--1865.

\bibitem[{Veli{\v{c}}kovi{\'c} et~al.(2018)Veli{\v{c}}kovi{\'c}, Cucurull, Casanova, Romero, Lio, and Bengio}]{velivckovic2018graph}
Petar Veli{\v{c}}kovi{\'c}, Guillem Cucurull, Arantxa Casanova, Adriana Romero, Pietro Lio, and Yoshua Bengio. 2018.
\newblock Graph attention networks.
\newblock In \emph{ICLR}.

\bibitem[{Wang et~al.(2020)Wang, Liu, Bi, Liu, He, Xu, and Yang}]{wang2020improving}
Jian Wang, Junhao Liu, Wei Bi, Xiaojiang Liu, Kejing He, Ruifeng Xu, and Min Yang. 2020.
\newblock Improving knowledge-aware dialogue generation via knowledge base question answering.
\newblock In \emph{AAAI}, pages 9169--9176.

\bibitem[{Wang et~al.(2022)Wang, Hou, Wang, Miao, Wu, Chen, Xia, Chi, Zhao, Liu et~al.}]{wang2022neural}
Yujing Wang, Yingyan Hou, Haonan Wang, Ziming Miao, Shibin Wu, Qi~Chen, Yuqing Xia, Chengmin Chi, Guoshuai Zhao, Zheng Liu, et~al. 2022.
\newblock A neural corpus indexer for document retrieval.
\newblock In \emph{NeurIPS}, pages 25600--25614.

\bibitem[{Wolf et~al.(2019)Wolf, Debut, Sanh, Chaumond, Delangue, Moi, Cistac, Rault, Louf, Funtowicz et~al.}]{wolf2019huggingface}
Thomas Wolf, Lysandre Debut, Victor Sanh, Julien Chaumond, Clement Delangue, Anthony Moi, Pierric Cistac, Tim Rault, R{\'e}mi Louf, Morgan Funtowicz, et~al. 2019.
\newblock Huggingface's transformers: State-of-the-art natural language processing.
\newblock \emph{arXiv:1910.03771}.

\bibitem[{Xu et~al.(2023)Xu, Sheng, Qi, Fu, Lin, Wang, and Zhou}]{xu-etal-2023-unsupervised}
Yi~Xu, Shuqian Sheng, Jiexing Qi, Luoyi Fu, Zhouhan Lin, Xinbing Wang, and Chenghu Zhou. 2023.
\newblock Unsupervised graph-text mutual conversion with a unified pretrained language model.
\newblock In \emph{ACL}, pages 5130--5144.

\bibitem[{Yasunaga et~al.(2021)Yasunaga, Ren, Bosselut, Liang, and Leskovec}]{yasunaga2021qa}
Michihiro Yasunaga, Hongyu Ren, Antoine Bosselut, Percy Liang, and Jure Leskovec. 2021.
\newblock {QA}-{GNN}: Reasoning with language models and knowledge graphs for question answering.
\newblock In \emph{NAACL-HLT}, pages 535--546.

\bibitem[{Young et~al.(2018)Young, Cambria, Chaturvedi, Zhou, Biswas, and Huang}]{young2017augmenting}
Tom Young, Erik Cambria, Iti Chaturvedi, Hao Zhou, Subham Biswas, and Minlie Huang. 2018.
\newblock Augmenting end-to-end dialogue systems with commonsense knowledge.
\newblock In \emph{AAAI}, pages 4970--4977.

\bibitem[{Yu et~al.(2022)Yu, Zhu, Yang, and Zeng}]{yu2022jaket}
Donghan Yu, Chenguang Zhu, Yiming Yang, and Michael Zeng. 2022.
\newblock Jaket: Joint pre-training of knowledge graph and language understanding.
\newblock In \emph{AAAI}, pages 11630--11638.

\bibitem[{Yu et~al.(2023)Yu, Iter, Wang, Xu, Ju, Sanyal, Zhu, Zeng, and Jiang}]{yu2023generate}
Wenhao Yu, Dan Iter, Shuohang Wang, Yichong Xu, Mingxuan Ju, Soumya Sanyal, Chenguang Zhu, Michael Zeng, and Meng Jiang. 2023.
\newblock Generate rather than retrieve: Large language models are strong context generators.
\newblock In \emph{ICLR}.

\bibitem[{Zhang et~al.(2020)Zhang, Liu, Xiong, and Liu}]{zhang2019grounded}
Houyu Zhang, Zhenghao Liu, Chenyan Xiong, and Zhiyuan Liu. 2020.
\newblock Grounded conversation generation as guided traverses in commonsense knowledge graphs.
\newblock In \emph{ACL}, pages 2031--2043.

\bibitem[{Zhang et~al.(2022{\natexlab{a}})Zhang, Zhang, Yu, Tang, Tang, Li, and Chen}]{zhang2022subgraph}
Jing Zhang, Xiaokang Zhang, Jifan Yu, Jian Tang, Jie Tang, Cuiping Li, and Hong Chen. 2022{\natexlab{a}}.
\newblock Subgraph retrieval enhanced model for multi-hop knowledge base question answering.
\newblock In \emph{ACL}, pages 5773--5784.

\bibitem[{Zhang et~al.(2022{\natexlab{b}})Zhang, Bosselut, Yasunaga, Ren, Liang, Manning, and Leskovec}]{zhang2022greaselm}
Xikun Zhang, Antoine Bosselut, Michihiro Yasunaga, Hongyu Ren, Percy Liang, Christopher~D Manning, and Jure Leskovec. 2022{\natexlab{b}}.
\newblock Grease{LM}: Graph {REAS}oning enhanced language models.
\newblock In \emph{ICLR}.

\bibitem[{Zhao et~al.(2020)Zhao, Wu, Tao, Xu, Zhao, and Yan}]{zhao2020low}
Xueliang Zhao, Wei Wu, Chongyang Tao, Can Xu, Dongyan Zhao, and Rui Yan. 2020.
\newblock Low-resource knowledge-grounded dialogue generation.
\newblock In \emph{ICLR}.

\bibitem[{Zhou et~al.(2021)Zhou, Huang, Liu, Chen, and Zhu}]{zhou2021earl}
Hao Zhou, Minlie Huang, Yong Liu, Wei Chen, and Xiaoyan Zhu. 2021.
\newblock {EARL}: Informative knowledge-grounded conversation generation with entity-agnostic representation learning.
\newblock In \emph{EMNLP}, pages 2383--2395.

\bibitem[{Zhou et~al.(2018)Zhou, Young, Huang, Zhao, Xu, and Zhu}]{zhou2018commonsense}
Hao Zhou, Tom Young, Minlie Huang, Haizhou Zhao, Jingfang Xu, and Xiaoyan Zhu. 2018.
\newblock Commonsense knowledge aware conversation generation with graph attention.
\newblock In \emph{IJCAI}, pages 4623--4629.

\bibitem[{Zhu et~al.(2024)Zhu, Chen, Wang, Gong, Yang, and Xie}]{zhu2024dyval}
Kaijie Zhu, Jiaao Chen, Jindong Wang, Neil~Zhenqiang Gong, Diyi Yang, and Xing Xie. 2024.
\newblock {DyVal}: Graph-informed dynamic evaluation of large language models.
\newblock In \emph{ICLR}.

\bibitem[{Zhu et~al.(2021)Zhu, Yang, Xu, Wang, Zhang, and Han}]{zhu2021transfer}
Qi~Zhu, Carl Yang, Yidan Xu, Haonan Wang, Chao Zhang, and Jiawei Han. 2021.
\newblock Transfer learning of graph neural networks with ego-graph information maximization.
\newblock In \emph{NeurIPS}, pages 1766--1779.

\end{thebibliography}
